\documentclass[lettersize,journal]{IEEEtran}
\usepackage{amsmath,amsfonts}
\usepackage{array}
\usepackage{textcomp}
\usepackage{stfloats}
\usepackage{url}
\usepackage{verbatim}
\usepackage{graphicx}
\usepackage{cite}

\usepackage{subfigure}
\usepackage{color}
\usepackage{multirow}
\usepackage{mathtools}
\usepackage{fancyvrb}

\usepackage{algorithmic}
\usepackage{algorithm}
\usepackage{epsfig}

\begin{document}

\title{OpenGV 2.0: Motion prior-assisted calibration and SLAM with vehicle-mounted surround-view systems}

\author{Kun Huang$^{1*}$, Yifu Wang$^{1*}$, Si'ao Zhang$^{1*}$, Zhirui Wang$^{2}$, Zhanpeng Ouyang$^{1}$, Zhenghua Yu$^{2}$, Laurent Kneip$^{1}$\\
$^{1}$ ShanghaiTech University, $^{2}$ Motovis Intelligent Technologies
\thanks{$^{*}$ equal contribution.}
\thanks{This paper is joint work between ShanghaiTech University and Motovis Intelligent Technologies. Contact: lkneip@shanghaitech.edu.cn}
}

\maketitle

\begin{abstract}
The present paper proposes optimization-based solutions to visual SLAM with a vehicle-mounted surround-view camera system. Owing to their original use-case, such systems often only contain a single camera facing into either direction and very limited overlap between fields of view. Our novelty consist of three optimization modules targeting at practical online calibration of exterior orientations from simple two-view geometry, reliable front-end initialization of relative displacements, and accurate back-end optimization using a continuous-time trajectory model. The commonality between the proposed modules is given by the fact that all three of them exploit motion priors that are related to the inherent non-holonomic characteristics of passenger vehicle motion. In contrast to prior related art, the proposed modules furthermore excel in terms of bypassing partial unobservabilities in the transformation variables that commonly occur for Ackermann-motion. As a further contribution, the modules are built into a novel surround-view camera SLAM system that specifically targets deployment on Ackermann vehicles operating in urban environments. All modules are studied in the context of in-depth ablation studies, and the practical validity of the entire framework is supported by a successful application to challenging, large-scale publicly available online datasets. Note that upon acceptance, the entire framework is scheduled for open-source release as part of an extension of the OpenGV library.
\end{abstract}

\begin{IEEEkeywords}
visual odometry, visual SLAM, bundle adjustment, continuous-time, autonomous driving, smart vehicles, OpenGV, surround-view, generalized camera, multi-perspective camera, calibration, Ackerman, non-holonomic
\end{IEEEkeywords}


\section{Introduction}

\IEEEPARstart{S}{mart} vehicle technology plays a key role in the 21st century’s automation revolution. Driven by the advent of high resolution 3D sensors and powerful parallel computing architectures running deep spatial AI networks, modern smart vehicles are able to drive thousands of kilometers in regular traffic conditions without the need of intervention~\cite{techradar_self_driving_cars}. As a result, key economic players of our world currently race towards the all-out introduction of autonomous vehicles as demonstrated by an increasing number of pilot commercial deployments. However, autonomous driving remains an expensive technology that is currently not yet accessible to common users. It is for this reason that the community maintains a high interest in the low-cost, exteroceptive perception system given by surround-view cameras. The latter is a close-to-market solution that is readily installed in today's passenger vehicles to provide the driver with a passive navigation assistance in the form of a bird-eye view to facilitate parking maneuvers.

\begin{figure}[t]
  \centering
  \includegraphics[width=0.82\columnwidth]{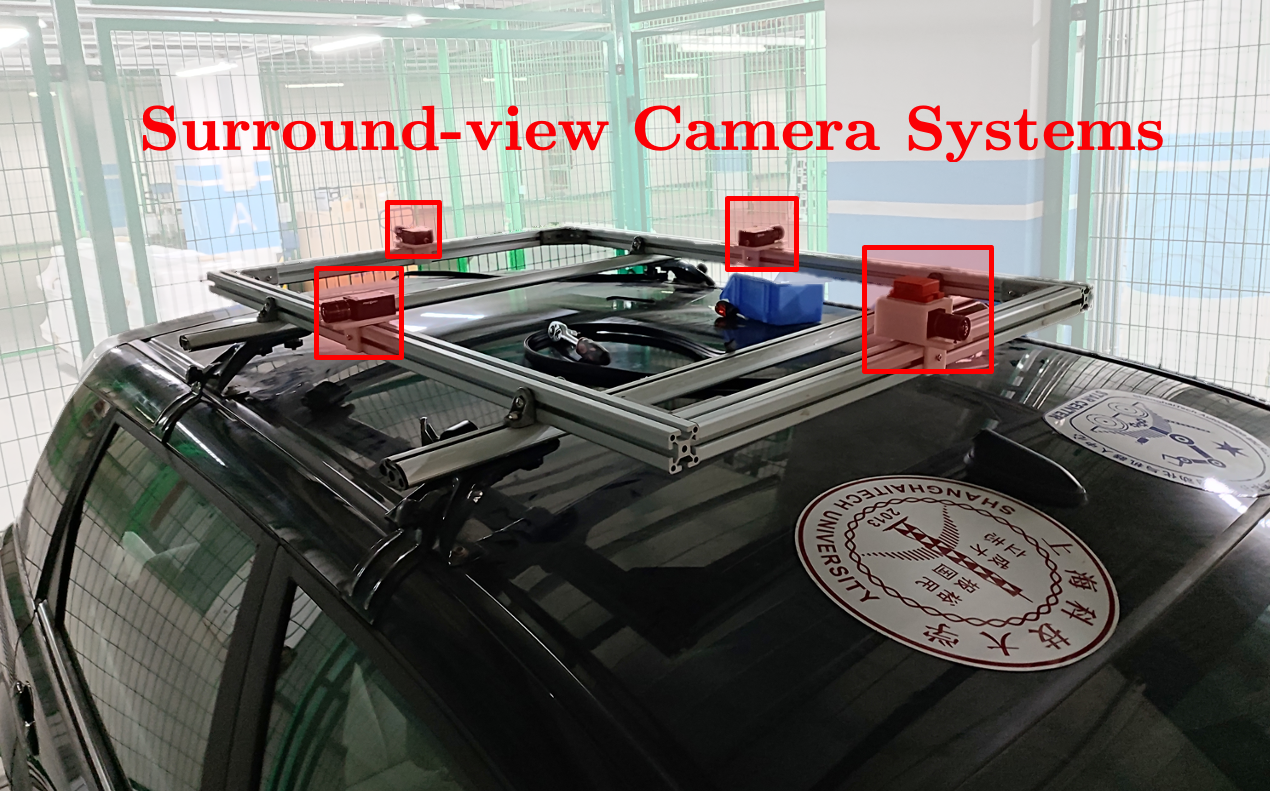}
  \caption{Example of a vehicle-mounted surround-view camera system as studied in this paper.}
  \label{fig:commonSystem}
\end{figure}

Solving localization and mapping problems with a surround-view camera system remains a challenge for a number of reasons. First, as illustrated in Figure~\ref{fig:commonSystem}, the system commonly has only four cameras which are pointing into different directions and have very limited overlap in their fields of view. Besides complicating the exploitation of stereo constraints for direct depth sensing, the lack of overlap already leads to difficulties during the initial extrinsic calibration of the system. Second, in order to achieve as complete as possible bird-eye views from only four cameras, the latter nonetheless have a large field of view and thus increased and varying distortions across the image plane. As a third challenge, the scale observability of the estimation is often weak given common degenerate motion conditions for Ackerman steering vehicles. Finally, the low-grade cameras employed in market-ready solutions often have disadvantageous properties such as high sensor noise and rolling shutter mechanisms.

The use of a surround-view camera system towards partial vehicle autonomy such as automated valet parking was pioneered by the project V-Charge~\cite{furgale13b}. The work has shown that such perception systems may provide sufficient information to perform online localization and mapping in unknown environments, and sub-sequently vision-based autonomous navigation in controlled, low-speed scenarios. The authors of the present paper have kept making contributions to the field, leading to the present full-stack compendium of algorithms that all share one common idea to further alleviate the above-listed challenges: the exploitation of motion priors. It is well understood that a car is a non-holonomic body for which the number of controllable degrees of freedom is less than the actual number of degrees of freedom of the vehicle. Even if restricting the vehicle pose to a 3D manifold (i.e. three degrees of freedom), the number of controllable degrees of freedom still remains less. In simple terms, given normal road conditions and moderate, slip-free motion, a car is unable to move side-ways, and its motion can be approximated locally by an arc of a circle within the local tangential plane.

The present paper summarizes the following contributions, which are all grounded on the idea of enforcing at least part of these constraints in order to facilitate, stabilize, or improve geometric calibration problems with non-overlapping surround-view camera systems:
\begin{itemize}
  \item We summarize the full stack of contributions made towards online calibration of camera-to-vehicle transformations, reliable vehicle-mounted multi-camera motion initialization, and large-scale continuous non-holonomic trajectory optimization using b-splines.
  \item We propose a complete system that integrates all modules and ultimately demonstrates highly reliable and accurate, real-time localization and mapping on large-scale publicly available urban benchmark sequences.
  \item We release an update to our existing open-source software package which extends the framework by the motion-prior constrained geometric surround-view solvers and optimizers presented in this work\footnote{Link will be provided upon acceptance}.
\end{itemize}
The paper is organized as follows. Section~\ref{sec:relatedWork} presents an extensive overview of the relevant literature as well as notations used throughout the paper. Section~\ref{sec:calibration} introduces the online method for flexible calibration of non-overlapping, vehicle-mounted surround-view camera systems. Section~\ref{sec:vo} introduces our stable frame-to-frame motion initialization technique that remains unaffected by motion degeneracies. Next, Section~\ref{sec:optimization} explains how to exactly impose non-holonomic vehicle motion constraints in back-end optimization. Section~\ref{sec:multivo} introduces the complete framework. Finally, Section~\ref{sec:results} shows all results obtained primarily on public benchmark datasets, before Section~\ref{sec:conclusion} concludes the work.

\textbf{Note that the present paper is an extension of our previous work~\cite{ouyang2020online, wang2020reliable, huang2021b}. The present work unifies and establishes the link between those prior methods in terms of being supported by vehicle-inherent motion priors, and furthermore introduces a complete surround-view camera SLAM system built on top of these modules. The framework is able to estimate complete trajectories in large-scale environments while supporting alternations between forward and backward motion as they would for example occur during parking maneuvers. The system is successfully applied to large-scale open-source sequences, and will be released upon acceptance.}


\section{Related work and preliminaries}
\label{sec:relatedWork}

This section reviews important related work according to the three discussed topics of motion-prior supported calibration, incremental motion estimation, and back-end optimization. We furthermore will see the notations used throughout the paper.

\subsection{Related work on extrinsic calibration}

With an aim to overcome the missing or reduced overlap of the fields of view of distinct cameras inside surround-view systems, the following extrinsic calibration approaches have been presented:
\begin{itemize}
  \item Infra-structure based calibration~\cite{ly2014extrinsic,choi2018automatic}: Multiple calibration targets with known relative transformations are fixed on the ground or to the walls of a calibration room. Concatenating target-to-camera transformations then leads to camera-to-camera transformations.
  \item Mirror-based calibration~\cite{kumar2008simple,lebraly2010flexible,takahashi2012new,long15}: A single calibration target is rendered visible in pairs of cameras by use of a sufficiently large mirror which is placed in front of one of the cameras. Calibration simultaneously solves for the mirror pose and two target-to-camera transformations.
  \item Loop-based calibration~\cite{furgale13,rehder16}: A single calibration target is moved around the vehicle and weak camera-to-camera transformations are derived from small pair-wise overlaps. The overall quality of the extrinsic transformations may be improved by imposing cycle constraints. 
  \item Hand-eye calibration~\cite{shiu1987calibration,tsai1989new,esquivel2007calibration,lebraly2011fast,pachtrachai2018chess,zhi2017simultaneous}: Cycle constraints are imposed on a concatenation of temporal camera displacements as well as extrinsic intra-camera transformation parameters. The method requires synchronized, accurate relative pose measurements from multiple cameras.
  \item SLAM-based calibration~\cite{carrera2011slam,heng2013camodocal,heng2014infrastructure}: Recovery of extrinsic parameters by aligning trajectories or performing extrinsic parameter-aware, large-scale bundle adjustment. Complete large-scale reconstruction is required for each camera, and the solution may be affected by drift.
\end{itemize}

The majority of extrinsic calibration methods do not attempt to identify camera-to-vehicle transformations, but only direct camera-to-camera transformations. However, in order to relate the motion of the camera array to the motion of the vehicle and exploit vehicle motion constraints, it is necessary to know exact camera-to-vehicle transformations~\cite{scaramuzza09,Huang19,lee13}, where the vehicle frame is defined such that non-holonomic motion constraints are easily expressed. Planar motion constraints for extrinsic calibration have already been explored by Lebraly et al.~\cite{lebraly10,lebraly2011fast}, who propose a closed-form solution for the camera-to-vehicle transformations. However, they fall back to bundle adjustment involving direct camera-to-camera transformations in the concluding optimization. Knorr et al.~\cite{knorr2013online} present a method that estimates planar homographies from ground plane observations in each frame, thus revealing the vertical direction in each camera. It is however difficult to reliably extract and match sufficient correspondences on the ground plane, which often does not provide sufficient texture. In this work, we propose an efficient online approach to identify and optimize two vehicle frame-related directional correspondences in each camera, and thereby find all camera-to-vehicle rotation matrices. The method does not require any calibration targets or large-scale SLAM solutions, and direct camera-to-camera transformations are naturally obtained as a by-product of the calculation.

Note that recently there has been a surge in extrinsic calibration approaches that apply novel rendering methods such as NerF~\cite{herau2024soac}. However, given that such methods are computationally demanding, often include lidar data in the process, and still leverage traditional solutions in order to initialize camera poses, the focus of our discussion here remains on geometric methods.

\subsection{Related work on motion initialization}

After successful calibration, the array of cameras in a surround-view system is commonly described by the generalized camera model~\cite{pless}. Generalized camera relative displacement solvers operating in six degrees-of-freedom have substantial computational complexity and solution multiplicity in the minimal case~\cite{stewenius05}, or require too many samples and linearizations in the non-minimal case, thus leading to unstable results under noise~\cite{li08}. A solution that factorizes the generalized relative pose problem as an iterative optimization over relative rotation is presented by Kneip and Li~\cite{kneip14}. The latter solver however degenerates in the case of non-holonomic motion, which is the primary target addressed by this work. An important summary contribution for generalized camera relative displacement is given by Kneip and Furgale~\cite{kneip14opengv}, who present the OpenGV framework with many robust solvers for solving geometric problems with generalized cameras. The library is the predecessor of the framework published in this work, and also includes many solvers to the standard relative pose problem (e.g. 8-point algorithm by Hartley~\cite{hartley97}, the 5-point algorithm by Stewenius and Nister~\cite{stewenius06}, or the eigensolver by Kneip and Lynen~\cite{kneip13}).

Scaramuzza et al.~\cite{scaramuzza09} introduce the first work that exploits the non-holonomic constraints of planar vehicles to parameterize the motion with only 1 feature correspondence. Huang et al.~\cite{Huang19} extend the approach to $n$ views. Lee et al.~\cite{lee13} furthermore apply the model to multi-perspective camera systems. While these methods are very robust, they still suffer from scale unobservability in the case of---for example---straight motion, and rely on the ideal assumption of a fixed steering angle; the centre of rotation as introduced by the Ackermann motion model is however a dynamic point for the majority of time during which a vehicle is taking a turn. Further related work is given by Lee et al.~\cite{lee14}, who look at the generalized relative pose problem with a known reference direction. This problem is highly related to the solution proposed in this work, as it only solves for a one-dimensional degree of freedom rotation. However, as shown in their work, the algorithm again potentially degenerates for planar vehicle motion, most notably if the relative rotation becomes identity. In this work, we introduce a robust generalized planar motion solver that simultaneously exploits the information from multiple cameras and is not affected by motion degeneracies. It may be regarded as an extension of the single-view solvers presented by Kneip and Lynen~\cite{kneip13}, or Booij et al.~\cite{Booij09}.

\subsection{Related work on back-end optimization and SLAM systems}

The third part of this work considers the improvement of visual SLAM solutions by including vehicle kinematics related constraints into the optimization. This practice is common in vision-based multi-sensor solutions that make additional use of odometers measuring the rotational velocity of each wheel. There have been EKF filter~\cite{wu17}, particle filter~\cite{yap11}, and optimization-based~\cite{quan18,kang2019vins} solutions, all relying on a drift-less planar motion model derived from a dual-drive or Ackermann steering platform. They perform relatively high-frequent integration of wheel odometry to come up with adequate priors on the relative displacement between subsequent views. Censi et al.~\cite{censi13} furthermore consider simultaneous extrinsic calibration between cameras and odometers. A closely related vehicle motion model that has also been used in filtering and optimization-based frameworks appears for skid-steering platforms~\cite{yi09,martinez17,lv17}. Another very closely related work is given by Zhang et al.~\cite{zhang19}, who continue to rely on a drift-less non-holonomic motion model, but extend the estimation to non-planar environments by introducing the motion-manifold and manifold-based integration of wheel odometry signals.

For pure vision-based solutions, the non-holonomic constraints need to be enforced purely by the model, which is more difficult. Long et al.~\cite{zong2017vehicle} and Li et al.~\cite{li18} propose the addition of regularization constraints into windowed optimization frameworks, which essentially penalize trajectory deviations from an approximate piece-wise circular arc model. From a purely geometric point of view, a drift-less, non-holonomic ground vehicle moves along smooth trajectories in space, and---more importantly---heads toward the vehicle motion direction. This motivates the use of the continuous-time trajectory model as proposed by Furgale et al.~\cite{furgale2015continuous}. While parametrizing a smooth vehicle trajectory, the representation and in particular its first-order differential is easily used to additionally enforce the vehicle heading to remain tangential to the trajectory.

The proposed back-end optimization simultaneously processes measurements emanating from all cameras. The advantages and challenges of monocular, stereo or multi camera visual-inertial SLAM have been discussed extensively in previous frameworks, such as~\cite{Qin2018VINSMONO,leutenegger2015keyframe,delmerico2018benchmark,van2023eqvio,qin2019general,Geneva2020OPENVINS,rosinol2020kimera,usenko2019visual,campos2021orb, zhang2021multi, wang2023mavis, kaveti2023design}. For a comprehensive survey, please refer to~\cite{Scaramuzza2016votut} and the latest research~\cite{huang2019review}. Here, we mainly focus on vision-only based solutions for multi-camera systems. \cite{urban2016multicol} extended ORB-SLAM2~\cite{murORB2} to multi-camera setups, supporting various rigidly coupled multi-camera systems. \cite{kuo2020redesign} introduced an adaptive SLAM system design for arbitrary multi-camera setups, requiring no sensor-specific tuning. Several works~\cite{furgale13b,wang2017scale,heng18,liu2018multi,wang2020reliable} focus on utilizing a surround-view camera system, often with multiple non-overlapping monocular cameras, or specializing on an application to motion estimation for ground vehicles. However, none of the listed frameworks is publicly available, and those proposing entire frameworks specialized on automotive applications~\cite{furgale13b,liu2018multi} are unable to operate in vision-only configuration.

\subsection{Notations and assumptions}

\begin{figure}[t]
  \centering
  \includegraphics[width=\columnwidth]{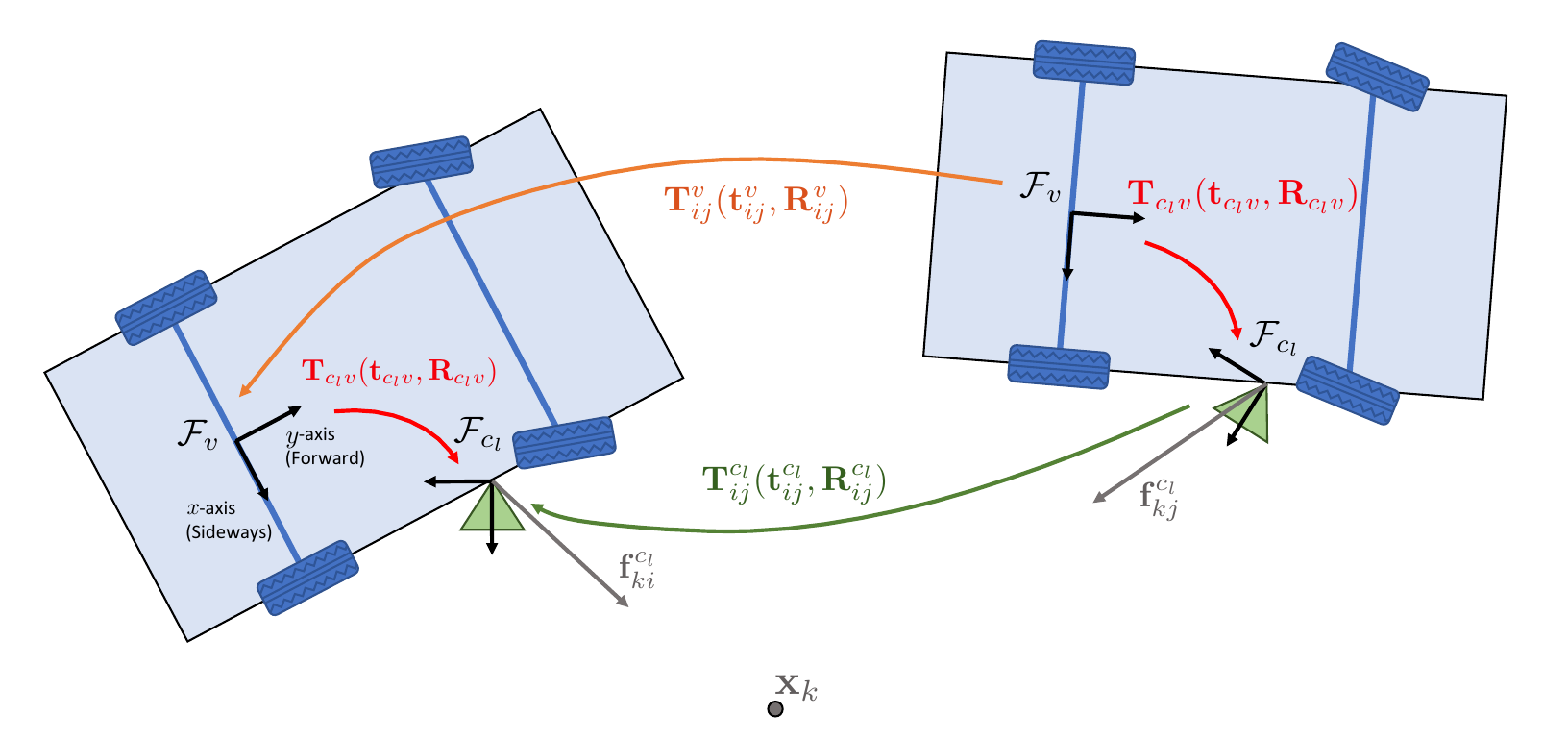}
  \caption{Notation of frames, landmarks, measurements, and extrinsic geometric transformation variables used in this work.}
  \label{fig:geometry}
\end{figure}

Though the number of cameras for all methods presented in this work is arbitrary, we assume the number to be four in most experiments. Furthermore, although the orientation and position of the cameras is theoretically arbitrary, we assume that the four cameras are mounted in the front, the back, and near the two side-mirrors. We furthermore assume that they are pointing forward, backward, and side-ways with approximately 90 degree angles between their principal axes. The arrangement is indicated in Figure~\ref{fig:commonSystem}. The cameras are furthermore assumed to be intrinsically calibrated. The projection from camera $c_l$ to the respective image frame is thus known and given by the abstract transformation function $\pi_{c_l}(\mathbf{x}^{c_l}) = \mathbf{u}^{c_l}$, where $\mathbf{x}^{c_l}$ is a Euclidean 3-vector defining a point in the camera frame, and $\mathbf{u}^{c_l}$ is a Euclidean 2-vector representing a point in the image plane. The cameras are assumed to have central projective geometry, and all transformation functions are assumed to be invertible. The transformation from image points into 3D bearing vectors defined in the camera frame is given by the known inverse mapping $\mathbf{f}^{c_l} = \pi_{c_l}^{-1}(\mathbf{u}^{c_l})$. Transformations in 3D are expressed by Euclidean transformation matrices $\mathbf{T}=\left[\begin{matrix} \mathbf{R} & \mathbf{t} \\ \mathbf{0} & 1 \end{matrix}\right]$, where $\mathbf{R}$ is a $3\times3$ rotation matrix and $\mathbf{t}$ is a $3\times1$ translation vector. Let $\mathcal{F}_{c_l}$ and $\mathcal{F}_v$ denote camera and vehicle reference frames. The extrinsic transformation from $\mathcal{F}_v$ to $\mathcal{F}_{c_l}$ is denoted by $\mathbf{T}_{c_lv} = \left[\begin{matrix} \mathbf{R}_{c_lv} & \mathbf{t}_{c_lv} \\ \mathbf{0} & 1 \end{matrix}\right]$ defined such that points defined in $\mathcal{F}_v$ are transformed to $\mathcal{F}_{c_l}$ according to the equation $\mathbf{x}^{c_l} = \mathbf{R}_{c_lv} \mathbf{x}^v + \mathbf{t}_{c_lv}$. Accordingly, $\mathbf{T}^v_{ij}$ and $\mathbf{T}^{c_l}_{ij}$ are the transformations that take 3D points from the vehicle or camera frame at time index $j$ to the ones at time index $i$, respectively. $\mathbf{T}^v_{i}$ takes points from the vehicle frame at time index $i$ to the global reference frame. Figure~\ref{fig:geometry} visualizes the geometric entities introduced here.

Note that the continuation mainly focuses on the geometric constraints built on top of sparse temporal correspondences. We assume that such feature extraction and matching techniques can be readily provided, and outline further details on our concrete implementation in the introduction of our complete visual SLAM framework in Section~\ref{sec:multivo}.


\section{Extrinsic calibration}
\label{sec:calibration}

Intrinsic calibration of cameras is a well-understood problem~\cite{zhang00,scaramuzza06toolbox}. Here we focus on a practical online approach to extrinsic calibration of surround-view camera systems with limited or no overlap between fields of view. The method returns camera-to-vehicle parameters, which are crucial for the application of motion-priors. However, note that the method is limited to the derivation of exterior orientations. This strategy is supported by the insight that camera positions are often well-defined by factory parameters as their tolerance is small enough to not have a significant impact on the accuracy of multi-perspective camera-based vehicle motion estimation. This is explained as follows:
\begin{itemize}
\item If the vehicle does not undergo any rotation, any point on the vehicle including the camera centre will be subject to the same translational displacement, and the rotation between different views will be identity. As a result, the measured displacement seen from the camera is going to be merely a function of extrinsic orientation, not position.
\item More general planar motion can be locally explained by the Ackermann steering model. Vehicle displacements are approximated by circular arcs, and the problem reduces to 1 DoF for any camera that is mounted in the vertical plane containing the non-steering axis~\cite{scaramuzza09}. Neither vertical nor lateral camera offsets will impact on either perceived camera rotation or direction of displacement. The radius of the ICR (i.e. the scale) is only observable by cameras that have a forward offset. Due to a typically small ratio between the forward offset and the depth of scene, the observability of scale remains however limited, and errors caused by wrong offset assumptions may easily be overshadowed by errors caused by general noise.
\end{itemize}
The use of factory parameters is further motivated by the fact that weak correlations between the camera mounting point and sensed camera displacements mean that the calibration of translational extrinsics from natural data would be challenging. Furthermore, note that translation parameters are typically stable over time. If accurate parameters are required, we therefore recommend the support of motion capture system-based offline calibration methods~\cite{kazik12,wang22calibration}. Here we focus on camera-to-vehicle rotations, which need post-factory calibration as well as continuous monitoring over time due to the influence of shocks and vibrations or temperature changes.

The proposed online calibration derives camera-to-vehicle rotations by identifying the vehicle frame-defining, motion-related directions in each camera, individually. We note that this calibration procedure does not require any calibration targets or large-scale SLAM solutions and that the camera-to-camera transformations are obtained as an immediate by-product of this derivation. We start by seeing the basic optimization constraint which is later complemented by camera-based measurements of the forward and upward directions.

\subsubsection{Basic optimization constraint}
We start from the epipolar incidence relationship. However, rather than employing it as a mere algebraic residual, we are considering a normalization of the error thus leading to the object space error (i.e. the orthogonal distance between two corresponding rays in space). Let $\mathbf{u}^{c_l}_{ki}$ and $\mathbf{u}_{kj}^{c_l}$ be corresponding image point measurements of the 3D point $\mathbf{x}_k$ in camera $c_l$ at time steps $i$ and $j$, respectively. Using $\mathbf{f}^{c_l}_{ki/j} = \pi_{c_l}^{-1}(\mathbf{u}^{c_l}_{ki/j})$, our optimization objective is given by
\begin{equation}
  \underset{\mathbf{R}^{c_l}_{ij},\mathbf{t}^{c_l}_{ij}}{\operatorname{argmin}} \sum_{k} \left( \frac{\mathbf{f}_{ki}^{c_lT} \lfloor \mathbf{t}^{c_l}_{ij} \rfloor_{\times} \mathbf{R}^{c_l}_{ij} \mathbf{f}_{kj}^{c_l}}{\| \lfloor \mathbf{f}_{ki}^{c_l} \rfloor_{\times} \mathbf{R}^{c_l}_{ij} \mathbf{f}_{kj}^{c_l}\|} \right)^2,
  \label{eq:relPoseEnergyGeometric}
\end{equation}
where $\mathbf{t}^{c_l}_{ij}$ is only optimized for direction and not for norm. The core reformulation consists of introducing the camera-to-vehicle rotation $\mathbf{R}_{c_lv}$ and to express the relative rotation of the camera frame as a function of the vehicle relative rotation, i.e.
\begin{eqnarray}
  \mathbf{R}^{c_l}_{ij} & = & \mathbf{R}_{c_lv}\mathbf{R}^v_{ij}\mathbf{R}_{c_lv}^T \nonumber\\
  \mathbf{R}^{c_l}_{ij} & = & \mathbf{R}_{c_lv}(\mathbf{R}^v_{i})^T\mathbf{R}^v_{j}\mathbf{R}_{c_lv}^T,
\end{eqnarray}
where $\mathbf{R}^v_{i}$ represents the absolute orientation of the vehicle frame at time $i$. Introducing the absolute vehicle orientation leads to the benefit of being able to reuse the same temporal orientation parameters for multiple cameras and multiple pairs of images for each camera. Supposing that we have $m$ frames and $d$ cameras in total, the joint geometric error is given by
\begin{eqnarray}
& & \underset{ \mathbf{R}^v_{1},\ldots,\mathbf{R}^v_{m}, \mathbf{R}_{c_1v}, \ldots, \mathbf{R}_{c_dv} }{\operatorname{argmin}} E_0 \nonumber \\
& = & \underset{ \mathbf{R}^v_{1},\ldots,\mathbf{R}^v_{m}, \mathbf{R}_{c_1v}, \ldots, \mathbf{R}_{c_dv} }{\operatorname{argmin}} \sum_{l=1}^{d} \sum_{\{ij\}}  \underset{\mathbf{t}^{c_l}_{ij}}{\operatorname{min}} \sum_k \epsilon_{lkij}^2 \text{ , where} \nonumber
\end{eqnarray}
\begin{equation}
\epsilon_{lkij} = \frac{\mathbf{f}_{ki}^{c_lT} \lfloor \mathbf{t}^{c_l}_{ij} \rfloor_{\times} \mathbf{R}_{c_lv}(\mathbf{R}^v_{i})^T\mathbf{R}^v_{j}\mathbf{R}_{c_lv}^T \mathbf{f}_{kj}^{c_l}}{\| \lfloor \mathbf{f}_{ki}^{c_l} \rfloor_{\times} \mathbf{R}_{c_lv}(\mathbf{R}^v_{i})^T\mathbf{R}^v_{j}\mathbf{R}_{c_lv}^T \mathbf{f}_{kj}^{c_l}\|}
\end{equation}
Note that---in a slight abuse of notation---the inner summation does not sum over all 3D points for all image pairs in each camera, but only over the visible points. Rotations are optimized as a function of a minimal, 3 DoF parametrization (e.g. Rodriguez vectors). Further note that, as outlined in~\cite{kneip13}, the internal minimization over the 2dof variable $\mathbf{t}^{c_l}_{ij}$ may be replaced by an eigenvalue-minimization problem that depends only on the rotation parameters. The algorithm only involves pair-wise constraints and rotation parameters, and is easier to realize than bundle adjustment. It does not optimize either 3D world points or the hardly observable vehicle-to-camera baselines $\mathbf{t}_{c_lv}$. Similar to~\cite{lebraly10}, our method has the advantage that the hand-eye calibration constraint is implicitly fulfilled by construction. However, it is also clear that the optimization has gauge freedom, and that the orientation of the vehicle frame $\mathcal{F}_v$ is thus far arbitrarily choosable. The vehicle frame orientation could be fixed to coincide with the orientation of one of the cameras, by which the exterior orientations would again become direct camera-to-camera transformations. However, the vehicle frame also has a clear definition based on motion related direction vectors that may be measured by each camera. In the following, we will introduce the resulting constraints.

\subsubsection{Constraints given by the forward direction}
Simply put, the columns of the exterior orientation matrix $\mathbf{R}_{c_lv}$ are the basis vectors of the vehicle frame expressed in the camera frame $\mathcal{F}_{c_l}$. The $x$-axis of the vehicle frame is pointing to the right, the $y$-axis to the front, and the $z$-axis upward.
Let us suppose that the vehicle is moving in the plane and currently not undergoing any rotation. The baselines of each camera are not going to contribute further to the displacement of the cameras, and every camera is going to measure the same displacement only expressed in a different frame, i.e. $\mathbf{t}^{c_l}_{ij}$. Given that we are on an Ackermann vehicle, this must be the forward direction, and $\mathbf{t}^{c_l}_{ij}$ therefore defines the second column of $\mathbf{R}_{c_lv}$ (i.e. the one corresponding to the forward direction). Again assuming that all $\mathbf{t}^{c_l}_{ij}$ have unit-norm, and using the notation $\lfloor \mathbf{R} \rfloor_{x}$ to denote a function that returns the $x$-th column of a rotation matrix, the constraint may be enforced by adding the energy term
\begin{equation}
   \underset{ \mathbf{R}_{c_1v}, \ldots, \mathbf{R}_{c_dv} }{\operatorname{argmin}} E_1 = \underset{ \mathbf{R}_{c_1v}, \ldots, \mathbf{R}_{c_dv} }{\operatorname{argmin}} \sum_{l=1}^{d} \sum_{\{ij\}} \| \mathbf{t}^{c_l}_{ij} - \lfloor \mathbf{R}_{c_lv} \rfloor_{2} \|^2
\end{equation}
Note that the pairs $\{i,j\}$ are selected if the angle of rotation is small enough. Given that the angle of rotation is independent of the reference frame, we simply detect forward displacements if $\|\mathbf{R}^{c_l}_{ij}-\mathbf{I}\|_{F}<\tau_1$, where $\tau$ is a small angular threshold (the Frobenius norm of $\mathbf{R}^{c_l}_{ij}-\mathbf{I}$ notably relates to the rotation angle of $\mathbf{R}^{c_l}_{ij}$).

\subsubsection{Constraints given by the upward direction}
Suppose that the angle of rotation now exceeds a certain threshold $\tau_2$, i.e. $\|\mathbf{R}^{c_l}_{ij}-\mathbf{I}\|_{F}>\tau_2$. Still assuming that the vehicle exerts planar motion, it is clear that the measured axis of rotation must correspond to the vertical, upward direction (i.e. the third column of $\mathbf{R}_{c_lv}$). Denoting $\{\mathbf{k}^{c_l}_{ij},\theta^{c_l}_{ij}\}$ to be the Euler axis-angle parameters of the rotation matrix $\mathbf{R}^{c_l}_{ij}$, we similarly arrive at the constraint
\begin{equation}
  \underset{ \mathbf{R}_{c_1v}, \ldots, \mathbf{R}_{c_dv} }{\operatorname{argmin}} E_2 = \underset{ \mathbf{R}_{c_1v}, \ldots, \mathbf{R}_{c_dv} }{\operatorname{argmin}} \sum_{l=1}^{d} \sum_{\{ij\}} \| \mathbf{k}^{c_l}_{ij} - \lfloor \mathbf{R}_{c_lv} \rfloor_{3} \|^2,
\end{equation}
where the pairs $\{i,j\}$ are chosen such that the minimum rotation angle constraint is satisfied.

\subsubsection{Including structural constraints}

\begin{figure}[t]
   \centering
   \includegraphics[width=0.8\linewidth]{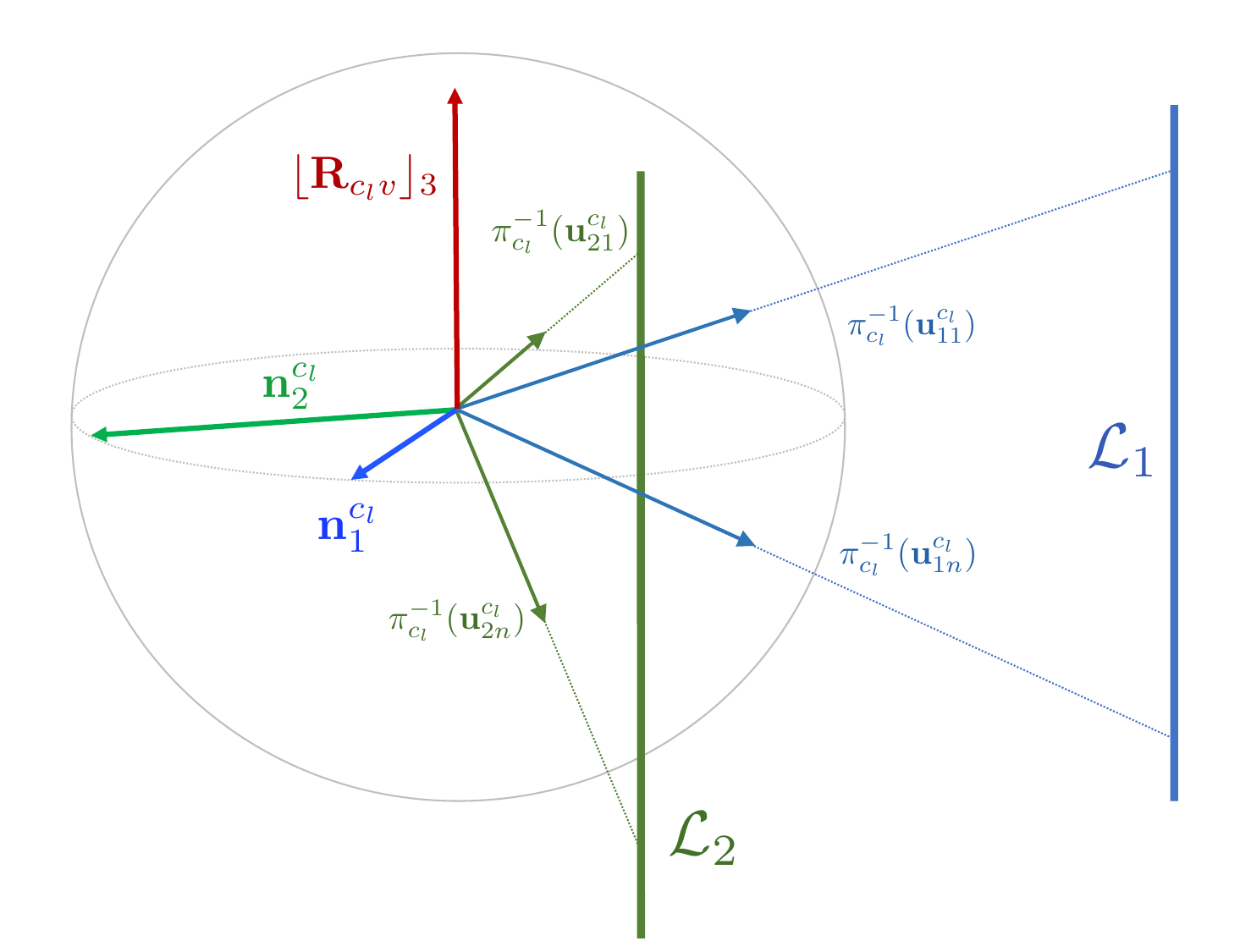}
   \centering
   \caption{The interpretation plane of a line $\mathcal{L}_k$ as defined by the normalized coordinates $\pi_{c_l}^{-1}(\mathbf{u}^{c_l}_{kij})$ of line measurements on a great circle.}
  \label{fig:line-model}
\end{figure}

Provided that the environment is man-made, the vertical direction may also be constrained by vertical lines. Let $\mathbf{u}^{c_l}_{kij},j=1,\ldots,n$ be point measurements believed to lie on a vertical line $\mathcal{L}_k$ in camera frame $c_l$ at time step $i$. We start by normalizing these points, which as a result come lying onto a great circle on the unit sphere around the camera. The normal vector $\mathbf{n}^{c_l}_{ki}$ of the interpretation plane passing through the camera centre and the line in 3D is finally recovered as the right-hand null-space of these bearing measurements
\begin{equation}
  \mathbf{n}^{c_l}_{ki} = \mathcal{NS} \left( \left[ \pi_{c_l}^{-1}(\mathbf{u}^{c_l}_{ki1}) \text{ } \ldots \text{ } \pi_{c_l}^{-1}(\mathbf{u}^{c_l}_{kin}) \right]^T \right).
\end{equation}
The vertical direction needs to be orthogonal to such a normal vector. If multiple vertical line measurements in multiple frames are given, we can further constraint the vertical direction of the vehicle by the energy minimization objective
\begin{equation}
  \underset{ \mathbf{R}_{c_1v}, \ldots, \mathbf{R}_{c_dv} }{\operatorname{argmin}} E_3 = \underset{ \mathbf{R}_{c_1v}, \ldots, \mathbf{R}_{c_dv} }{\operatorname{argmin}} \sum_{l=1}^{d} \sum_i \sum_k \| (\mathbf{n}^{c_l}_{ki})^{T} \cdot \lfloor \mathbf{R}_{c_lv} \rfloor_{3} \|^2
\end{equation}
The geometry of this matter is illustrated in Figure \ref{fig:line-model}.

\subsubsection{Overall optimisation objective}
To conclude, we swap the $L2$ norms for robust Huber norms and optimise the exterior vehicle-to-camera orientation parameters using the combined objective
\begin{equation}
  \underset{\mathbf{R}^v_{1}, \ldots,\mathbf{R}^v_{m}, \mathbf{R}_{c_1v}, \ldots, \mathbf{R}_{c_dv}}{\operatorname{argmin}} E_0 + \lambda_1 E_1 + \lambda_2 E_2 + \lambda_3 E_3.
\end{equation}
The weighting factors $\lambda_1$, $\lambda_2$, and $\lambda_3$ generate a trade-off between the various error terms. In particular, $\lambda_1$, $\lambda_2$, and $\lambda_3$ are chosen dynamically to for example balance the number of vertical line measurements against the number of points in each image. Note that as before, all rotations are parametrized minimally using 3 DoF Rodriguez vectors. Furthermore, note that the internal minimization over $\mathbf{t}^{c_l}_{ij}$ is done over a single degree of freedom, which is the angle of the translation direction vector in the horizontal plane as defined by $\lfloor \mathbf{R}_{c_lv} \rfloor_{3}$ in each camera.


\section{Stable motion initialization}
\label{sec:vo}

In analogy to \eqref{eq:relPoseEnergyGeometric}, the non-normalized (i.e. algebraic) sum-of-squared epipolar errors between a pair of views $i$ and $j$ from camera $c_l$ is given by
\begin{equation}
  \underset{\mathbf{t}_{ij}^{c_l},\mathbf{R}_{ij}^{c_l}}{\operatorname{argmin}}\sum_k ( \mathbf{f}_{ki}^{{c_l}T} \lfloor \mathbf{t}_{ij}^{c_l} \rfloor _\times \mathbf{R}_{ij}^{c_l} \mathbf{f}_{kj}^{c_l} )^2\,,
\end{equation}
where $\mathbf{f}_{ki}^{c_l}$ and $\mathbf{f}_{kj}^{c_l}$ still represent the unit bearing vectors pointing at the same 3D world point $\mathbf{x}_k$ from the $i$-th and $j$-th viewpoints of camera $c_l$, respectively, and $\mathbf{t}_{ij}^{c_l}$ and $\mathbf{R}_{ij}^{c_l}$ denote the respective euclidean transformation parameters. Note that if only a single relative pose between a pair of views is optimized, the norm of $\mathbf{t}_{ij}^{c_l}$ remains unconstrained. As illustrated in~\cite{kneip13}, we can apply the scalar triple product rule to the algebraic incidence relationship, and---by defining the epipolar plane normal vector as
\begin{equation}
    \mathbf{n}_{kij}^{c_l} = \mathbf{f}_{ki}^{c_l} \times \mathbf{R}{_{ij}^{c_l}}\mathbf{f}{_{kj}^{c_l}}\,,
\label{eq:1}
\end{equation}
---we easily arrive at the following modified objective for the algebraic energy minimization
\begin{equation}
\underset{\mathbf{t}_{ij}^{c_l},\mathbf{R}_{ij}^{c_l}}{\operatorname{argmin}} \text{ }\mathbf{t}_{ij}^{{c_l}T} \left(\sum_k  \mathbf{n}{_{kij}^{c_l}} \mathbf{n}{_{kij}^{{c_l}T}} \right) \mathbf{t}_{ij}^{c_l}.
\label{eq:eigenvalue}
\end{equation}
This objective is simple to solve by a minimization of the smallest eigenvalue of the matrix $\sum_k  \mathbf{n}{_{kij}^{c_l}} \mathbf{n}{_{kij}^{{c_l}T}}$, which only depends on $\mathbf{R}{_{ij}^{c_l}}$. Furthermore, as illustrated in~\cite{kneip14}, the matrix can be augmented to a $4\times 4$ matrix to solve the generalized case. However, as further explained in~\cite{kneip14}, the eigenvalue minimization objective is not well suited for multi-camera arrays, as the identity rotation always leads to a zero energy, even in the case of non-zero-angle rotations. As further illustrated in~\cite{li08} and~\cite{lee14}, linear formulations using the generalized essential matrix are affected by motion degeneracies, which unfortunately include pure translations and displacements along circular arcs. The fact that most displacements on a car are at least close to these types of motion means that linear solutions are also practically inapplicable. In the following, we will see a solution that is able to handle all kinds of planar motion.

Our proposed approach for estimating general planar motion with non-overlapping multi-camera systems consists of a parallel evaluation of the epipolar geometry for each individual camera as a univariate multi-eigenvalue problem. Let $\mathbf{n}{_{kij}^{c_l}}$ still be an epipolar plane normal vector as defined in \eqref{eq:1}. In the case of a calibrated multi-camera system, we can use the known extrinsic rotations $\mathbf{R}_{vc_l}$ to rotate all observed unit bearing vectors of camera $c_l$ into a frame that is still centered at camera $c_l$, but has similar orientation than the local vehicle frame. We can thus obtain an alternative multi-camera system in which all cameras simply have the same orientation than the local body frame. It can be easily observed that all the new normal vectors expressed in the body frame and given by
\begin{equation}
    \mathbf{n}_{kij}^{vc_l} = (\mathbf{R}_{vc_l}\mathbf{f}_{ki}^{c_l}) \times \mathbf{R}_{ij}^v(\mathbf{R}_{vc_l}\mathbf{f}_{kj}^{c_l})
\label{eq:2}
\end{equation}
still span a plane that is orthogonal to the translation $\mathbf{t}^{vc_l}_{ij}$ (even if the latter---which represents the translation of camera $c_l$ between views $i$ and $j$---is now expressed in the vehicle frame $v$). Similar to~\cite{kneip13,kneip14}, our target remains a solution to the relative displacement that depends only on $\mathbf{R}_{ij}^v$. 

Note that in the following, we drop indices $ij$ as the entire problem is essentially a two-view problem, and we furthermore move back to multiple cameras $c_l, l=1,\ldots,d$. In order to enforce all normal vectors of each camera to obey the coplanarity condition, the basic approach consists of stacking the normal vectors from each camera into the matrix $\mathbf{N}^{vc_l} = [ \mathbf{n}{_1^{vc_l}}\quad...\quad\mathbf{n}{_k^{vc_l}}]^{T}$ such that $\mathbf{N}^{vc_l} \mathbf{t}^{vc_l} = 0 $. Thus, the relative rotation $\mathbf{R}^{v}$ can be derived by jointly minimizing the smallest eigenvalue of the matrices $\mathbf{M}^{vc_l} = \mathbf{N}^{vc_l}\mathbf{N}^{vc_lT}$ from each camera. 
If $\lambda_{\mathbf{M}^{vc_l},min}$ denotes the smallest eigenvalue of $\mathbf{M}^{vc_l}$, our final objective becomes
\begin{equation}
    \mathbf{R}^v = \underset{\mathbf{R}^v}{\operatorname{argmin}} \sum_{l=1}^d (\lambda_{\mathbf{M}^{vc_l},min})^2\text{, where}
    \label{eq:basic}
\end{equation}
\begin{equation}
    \mathbf{M}^{vc_l} = \sum_{k} (\mathbf{R}_{vc_l}\mathbf{f}_{ki}^{c_l} \times \mathbf{R}^v\mathbf{R}_{vc_l}\mathbf{f}_{kj}^{c_l})(\mathbf{R}_{vc_l}\mathbf{f}_{ki}^{c_l} \times \mathbf{R}^v\mathbf{R}_{vc_l}\mathbf{f}_{kj}^{c_l})^T.
\end{equation}
It is important to realize that this objective is different from \eqref{eq:eigenvalue} in~\cite{kneip13}. Unless proceeding to a generalization as presented in~\cite{kneip14}, it is not possible to add all normal vectors to one co-planarity condition as each camera has a potentially different translation vector, and therefore defines a different plane for its epipolar plane normal vectors. However, the rotation is the same for each camera, and---owing to the fact that the eigenvalue formulation only depends on the relative rotation---we may still jointly minimize all objectives.

In the following, we concentrate on the case of planar motion, for which the relative rotation has only a single degree of freedom. We choose the Cayley~\cite{cayley46} parameters $ \mathbf{v} = [0\ 0\ z]^T $ to represent the rotation $\mathbf{R}^{v}$, the latter being given as
\begin{equation}
    \mathbf{R}^{v} = 2(\mathbf{v}\mathbf{v}^T - \lfloor\mathbf{v}\rfloor_{\times} ) + (1 - \mathbf{v}^T\mathbf{v})\mathbf{I}.
\end{equation}
Note that we omit the scale factor as it equally affects all terms in all energies. By using the uni-variate formulation, our objective turns into a very efficient non-linear optimization over a single parameter only. Note that the direction of each camera's relative translation can be recovered by looking at the eigenvector that corresponds to the smallest eigenvalue. As shown in the continuation, they can be further used to compute the scaled relative translation between the two viewpoints once the relative rotation $\mathbf{R}^v$ has been found. Similar to~\cite{kneip13}, the non-linear problem can be efficiently solved by implementing a Levenberg-Marquardt scheme. We sample the space of $z$ to initialize the local search.

\subsubsection{Object-space error refinement}
\label{sec:obj-space}
For a perspective camera, a purely translational displacement that is parallel to the image plane can cause a very similar disparity than a pure rotation around an orthogonal axis in the image plane, and vice-versa. On vehicles, the hereby described \textit{rotation-translation ambiguity} is furthermore amplified by sideways looking, fronto-parallel cameras, especially if they have a very limited field of view (FoV). The separation into 4 eigenvalue problems that are solved in parallel naturally raises the question how this affects the algorithm's ability to deal with such ambiguities. Though the algebraic solution is not geometrically or statistically meaningful, it generally leads to satisfying results at a low computational cost. However, the rotation-translation ambiguity can easily lead to local minima in the algebraic objective error in the above described case. In an aim to solve this problem, we go back to the object-space error-based objective as an efficient iterative refinement step. The object-space error is defined as the distance between the rays defined by $\mathbf{f}_{ki}^{c_l}$ and $\mathbf{f}_{kj}^{c_l}$. Starting from the definition of the distance between two skew lines~\cite{Gellert89}, we derive the geometric object-space error for a single correspondence to be
\begin{equation}
    d_k^{c_l} = \frac{ (\mathbf{R}_{vc_l}\mathbf{f}_{ki}^{c_l} \times \mathbf{R}^{v}\mathbf{R}_{vc_l}\mathbf{f}{_{kj}^{c_l}})\cdot\vec{\mathbf{t}^{vc_l}} }{ \|\mathbf{R}_{vc_l}\mathbf{f}_{ki}^{c_l} \times \mathbf{R}^{v}\mathbf{R}_{vc_l}\mathbf{f}{_{kj}^{c_l}}\|}.
\end{equation}
%
%
$\vec{\mathbf{t}^{vc_l}}$ represents the direction of the relative translation of camera $c_l$ expressed in the vehicle frame $v$. The optimization problem is finally given as
\begin{equation}
    \{\mathbf{R}^{v},\vec{\mathbf{t}^{vc_l}}\} = \underset{\mathbf{R}^v,\vec{\mathbf{t}^{vc_l}}}{\operatorname{argmin}} \sum_{k} (d_k^{c_l})^2.
\end{equation}
It is easy to see that the same objective can again be minimized by solving the iteratively reweighted eigenvalue-minimization problem
\begin{eqnarray}
    \mathbf{R}^v & = & \underset{\mathbf{R}^v}{\operatorname{argmin}} \sum_{l=1}^d (\lambda_{\tilde{\mathbf{M}}^{vc_l},min})^2 \text{, where}
    \label{eq:basic_optim} \\
    \tilde{\mathbf{M}}^{vc_l} & = & \sum_{k}
    \frac{(\mathbf{n}_k^{vc_l})(\mathbf{n}_k^{vc_l})^T}
    {\|\mathbf{n}_k^{vc_l}\|_2^2}.
\end{eqnarray}
%
%
The proposed object-space error minimization strategy still depends only on the relative rotation $\mathbf{R}^v$, meaning a one-dimensional optimization space in the case of planar motion. We confirmed through a series of simulation experiments that the minimization of the object-space error is more stable for different FoVs than the algebraic objective. It can effectively avoid wrong minima caused by rotation-translation ambiguity. At the same time, the computational complexity of the presented object-space error minimization objective is significantly lower than the one of standard two-view bundle adjustment. The latter not only optimizes over both rotation and translation parameters, but---if using the classical reprojection error---also over the 3D coordinates of each landmark. A dedicated experiment comparing the two iterative refinement alternatives is presented in Section \ref{sec:objVSBA}.

\subsubsection{Recovery of relative translation}
In order to recover the translation in absolute scale, we start by formulating the hand-eye calibration constraint for camera $c_l$ inside the multi-camera system~\cite{hartley04}:
\begin{equation}
\left\{
  \begin{aligned}
  \mathbf{t}^{v}  &=  \mathbf{t}_{vc_l} + \mathbf{R}_{vc_l}\mathbf{t}^{c_l} - \mathbf{R}_{vc_l}\mathbf{R}^{c_l}\mathbf{R}_{vc_l}^{T}\mathbf{t}_{vc_l}\\
  \mathbf{R}^{v}  &=  \mathbf{R}_{vc_l}\mathbf{R}^{c_l}\mathbf{R}_{vc_l}^{T}\\
  \end{aligned}
\right.
\label{eq:rigconstraint}
\end{equation}
As mentioned before, we can compensate for each camera's extrinsic rotation $\mathbf{R}_{vc_l}$, and thus obtain the simpler constraint
\begin{equation}
  \mathbf{t}^{v} =  \mathbf{t}_{vc_l} + \mathbf{t}^{vc_l} - \mathbf{R}^v\mathbf{t}_{vc_l}.
  \label{eq:finalconstraint}
\end{equation}
For each relative translation $\mathbf{t}^{vc_l} = \lambda_l \cdot\vec{\mathbf{t}^{vc_l}}$, directions $\vec{\mathbf{t}^{vc_l}}$ can be easily computed by composing $\tilde{\mathbf{M}}^{vc_l}$ and deriving the eigenvector corresponding to the optimized $\mathbf{R}^v$. All pair-wise constraints in the form of \eqref{eq:finalconstraint} can now be grouped into a linear problem $\mathbf{A}\mathbf{x} = \mathbf{b}$, where

\begin{small}
\begin{equation}
\mathbf{A} = \left[
  \begin{array}{cccc}
    \vec{\mathbf{t}^{vc_1}} &   &  & -\mathbf{I} \\
                   &...&  &...  \\
    & & \vec{\mathbf{t}^{vc_d}} & -\mathbf{I}
  \end{array}
\right], \text{ }
\mathbf{b} = \left[
  \begin{array}{c}   
      (\mathbf{R}^v - \mathbf{I})\mathbf{t}_{vc_1} \\
        ...\\
      (\mathbf{R}^v - \mathbf{I})\mathbf{t}_{vc_d}
  \end{array}
\right],
\end{equation}
\end{small}

\noindent and $\mathbf{x} = \left[\lambda_1 \text{ } \ldots \lambda_d \text{ } \mathbf{t}^{vT} \right]^T$. $\mathbf{A}$ and $\mathbf{b}$ can be computed from the known extrinsics and the relative rotation $\mathbf{R}^v$, whereas $\mathbf{x}$ contains all unknowns. The non-homogeneous linear problem $\mathbf{A}\mathbf{x}=\mathbf{b}$ can be solved by a standard technique such as singular value decomposition (SVD). Note that the system shows an obvious characteristic of non-overlapping multi-camera arrays, namely that metric scale remains unobservable if $\mathbf{R}^v=\mathbf{I}$. The rotation however remains computable.

\section{Back-end optimization}
\label{sec:optimization}

We now proceed to the final tool of the motion-aware geometric solvers presented in this work, which is the back-end optimizer. We start with seeing some preliminaries required for the later introduction of all vehicle motion model-aware objective functions for the back-end optimization. Note that the latter are generally formulated as a function of the Gold-standard error (i.e. the geometric reprojection error).

\subsection{B-splines}

Continuous-time parametrizations have shown great value in motion estimation when dealing with smooth trajectories or temporally dense sampling sensors. There are various alternatives for the basis functions, such as FFTs, discrete cosine transforms, polynomial kernels, or B\'ezier splines. In this paper, we will use the efficient and smooth B-spline parametrization~\cite{piegl2012nurbs} as promoted by Furgale et al.~\cite{furgale2015continuous}. We represent smooth motion with a $p$-th degree B-spline curve
\vspace{-0.1cm}
\begin{equation}
        \mathbf{c}(t) = \sum_{i=0}^{n} N_{i,p}(t) \mathbf{p}_i ,\qquad a\leq t\leq b,
\end{equation}
where $t$ is the continuous-time parameter, $\{\mathbf{p}_i\}$ are the $n+1$ control points that control the smooth trajectory shape, and $\{N_{i,p}(t)\}$ are the $n+1$ $p$th-degree B-spline basis functions. Note that the form of a B-spline and the basis functions are generally fixed, and the shape of the curve is influenced by the control points only. Trajectory splines are initialized from a set of discrete vehicle poses for each image and the image time-stamps\footnote{Note that we use cubic B-splines which ensures two times continuous differentiability. }. We use the spline curve approximation algorithm presented in Piegl and Tiller~\cite{piegl2012nurbs} for the initialization. The reader is invited to see more detailed foundations of B-splines and an example application in~\cite{piegl2012nurbs} and~\cite{furgale2015continuous}.

\subsection{Non-holonomic motion}
\label{subsec:nonholonomic}

Ground vehicles commonly have a non-steering two-wheel axis, which causes the motion to be non-holonomic. This kinematic constraint is reflected in the Ackermann steering model. Infinitesimal motion is a rotation about an Instantaneous Centre of Rotation (ICR) which lies on the extended non-steering two-wheel axis. In other words, the instantaneous heading of the vehicle is parallel to its velocity. The constraint has already been exploited in purely vision-based algorithms, however only based on the approximation of a piece-wise constant steering angle:

\begin{itemize}
    \item Front-end: Scaramuzza et al.~\cite{scaramuzza09} approximate the motion to be on a plane and the platform to have a locally constant steering angle. The trajectory between subsequent views is hence approximated by an arc of a circle, and the heading remains tangential to this arc. A minimal parameterization of the motion is given by the inscribed arc-angle $\theta$ as well as the radius of this circle $r$, and both the relative rotation and translation are expressed as functions of these parameters.
    \item Back-end: As proposed by Peng et al.~\cite{peng2019articulated}, relative rotations $\mathbf{R}^v$ and translations $\mathbf{t}^v$ under the approximation of a piece-wise constant steering angle or circular arc model need to satisfy the constraint
    \vspace{-0.1cm}
    \begin{equation}
        \left(\left(\mathbf{I}+\mathbf{R}^v\right)\begin{bmatrix}0&1&0\end{bmatrix}^T\right) \times \mathbf{t}^v = 0,
        \label{eq:rtconstraint}
    \end{equation}
    which can be added as a regularization term in a common bundle adjustment framework. We call this the \textit{R-t constraint}.
\end{itemize}

The above models are only an approximation of the original infinitesimal constraints on the velocity and the position of the ICR. In the continuation, we will introduce the use of continuous-time parametrizations to continuously enforce identity between the body's velocity direction $\frac{\mathbf{v}^v}{\|\mathbf{v}^v\|}$ and the vehicle's forward axis $y_v$, which is the original infinitesimal constraint. We call this the \textit{R-v constraint}.

\subsection{Optimization of non-holonomic trajectories}\label{sec:theory}

We enforce the R-v constraint by using a spline to represent the non-holonomic vehicle trajectory in continuous time. The first-order differential of the spline gives us the instantaneous velocity of the vehicle, which we can then use to either directly express the vehicle heading (i.e. as a hard constraint), or otherwise form a regularization term on the vehicle orientation (i.e. as a soft constraint). Imposing the kinematic constraints as a hard or soft constraint may impact on both accuracy and computational performance, which is why we introduce and compare multiple formulations starting from conventional bundle adjustment.

\begin{figure*}[htb]
\vspace{0.2cm}
\centering
\subfigure[]{\label{fig:CBA}\includegraphics[width=0.15\linewidth]{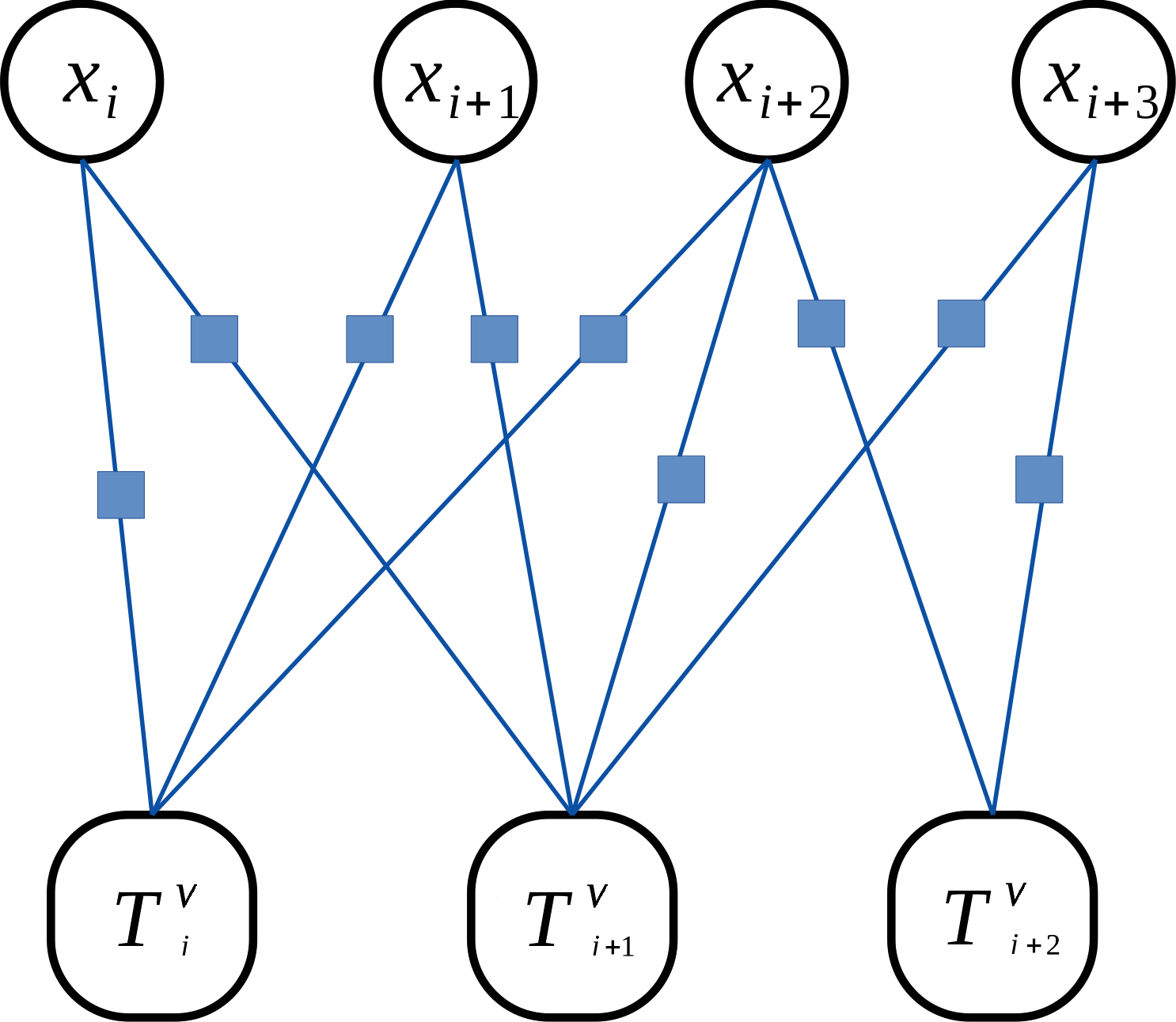}}
\subfigure[]{\label{fig:CBA_Rt}\includegraphics[width=0.18\linewidth]{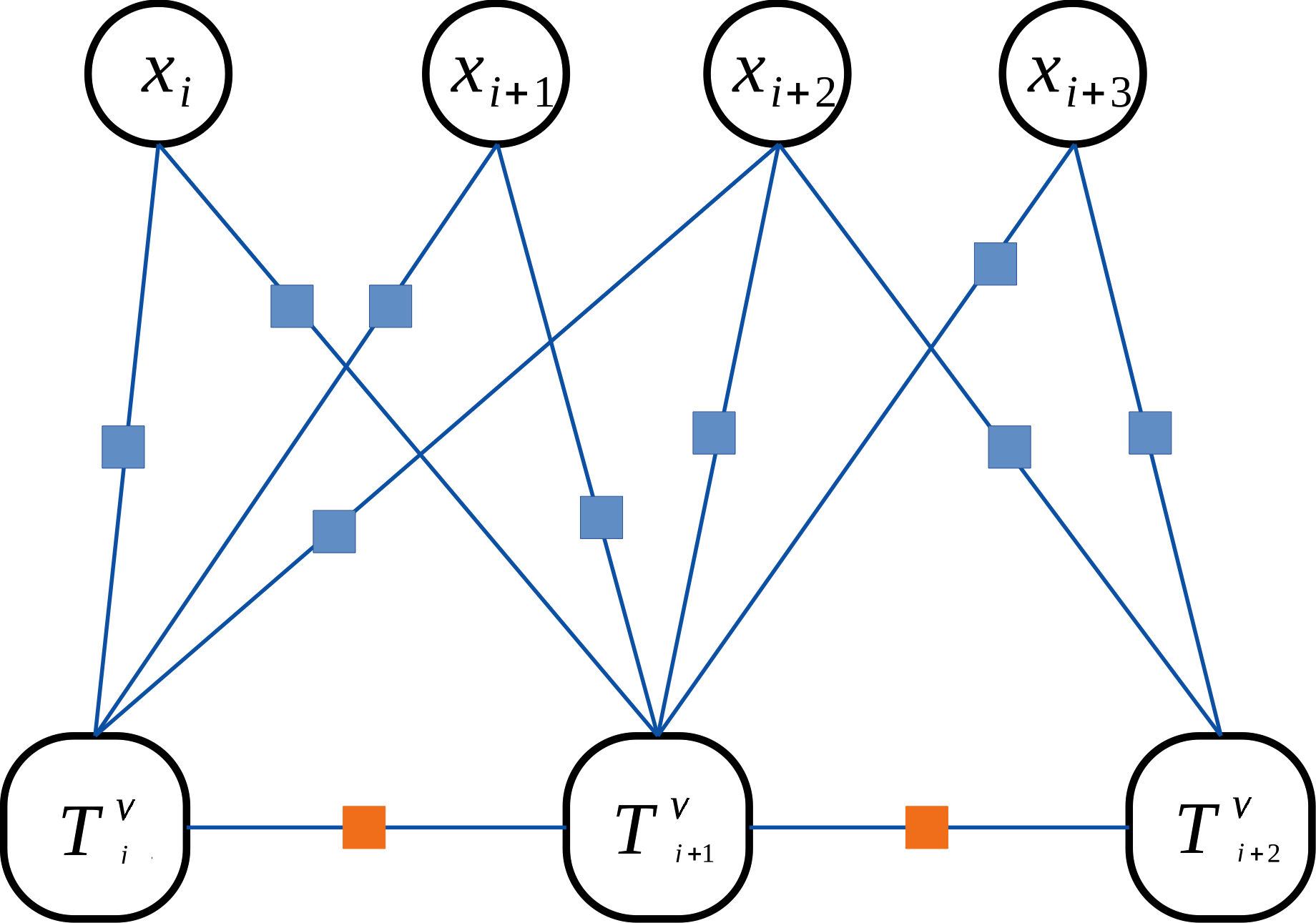}}
\subfigure[]{\label{fig:CBA_Rv}\includegraphics[width=0.22\linewidth]{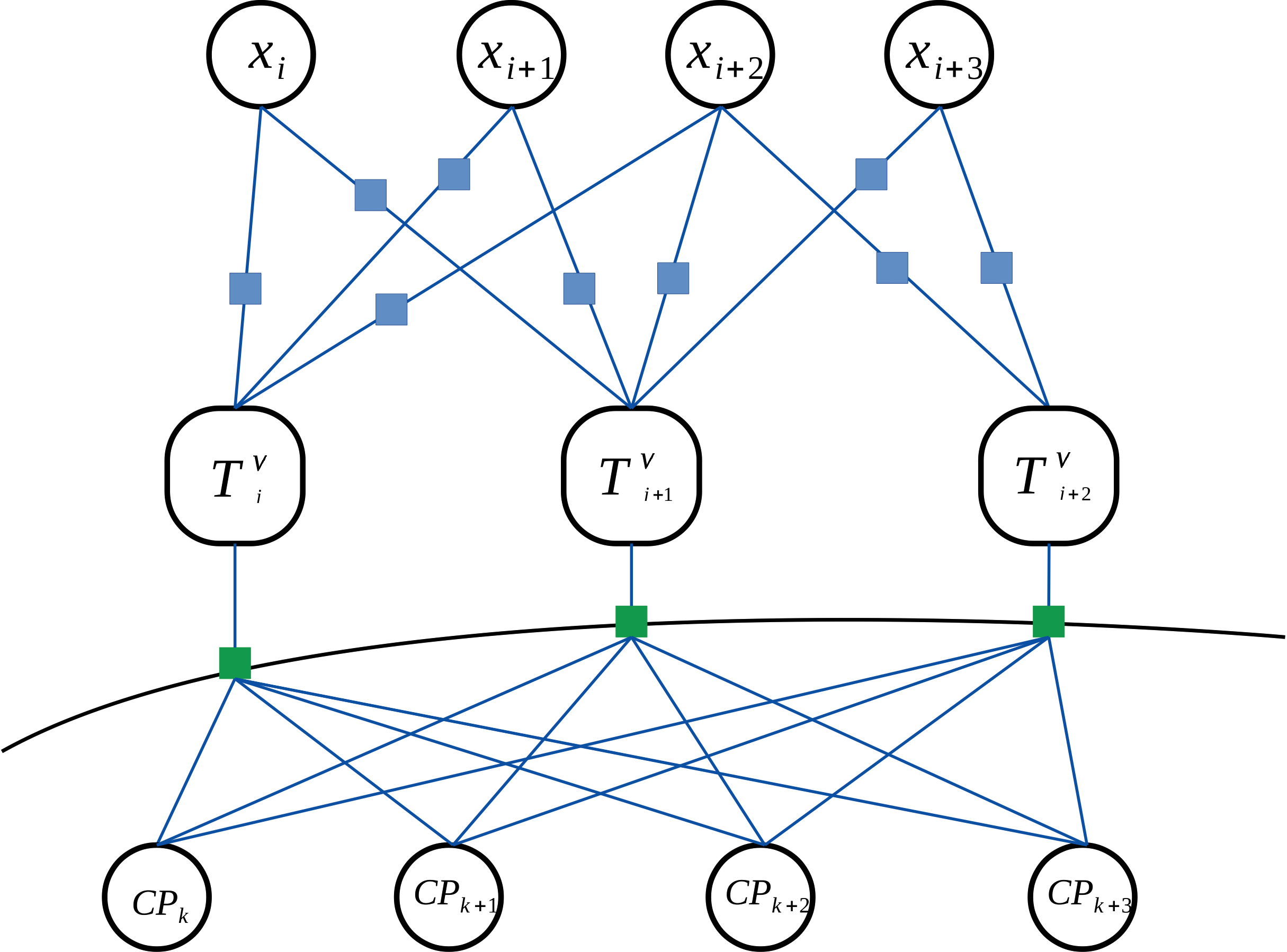}}
\subfigure[]{\label{fig:SSBA_Rv}\includegraphics[width=0.19\linewidth]{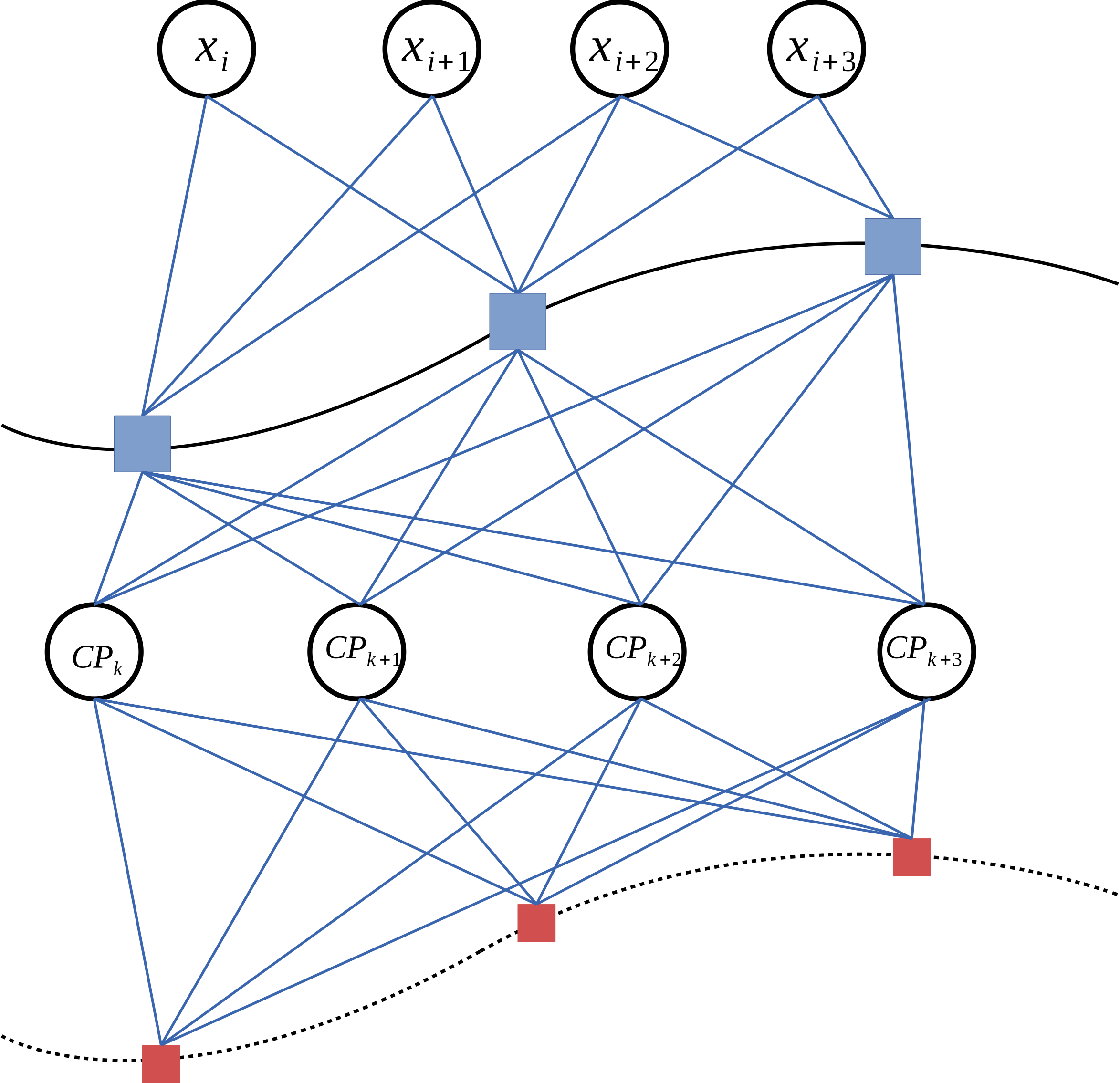}}
\subfigure[]{\label{fig:FSBA}\includegraphics[width=0.20\linewidth]{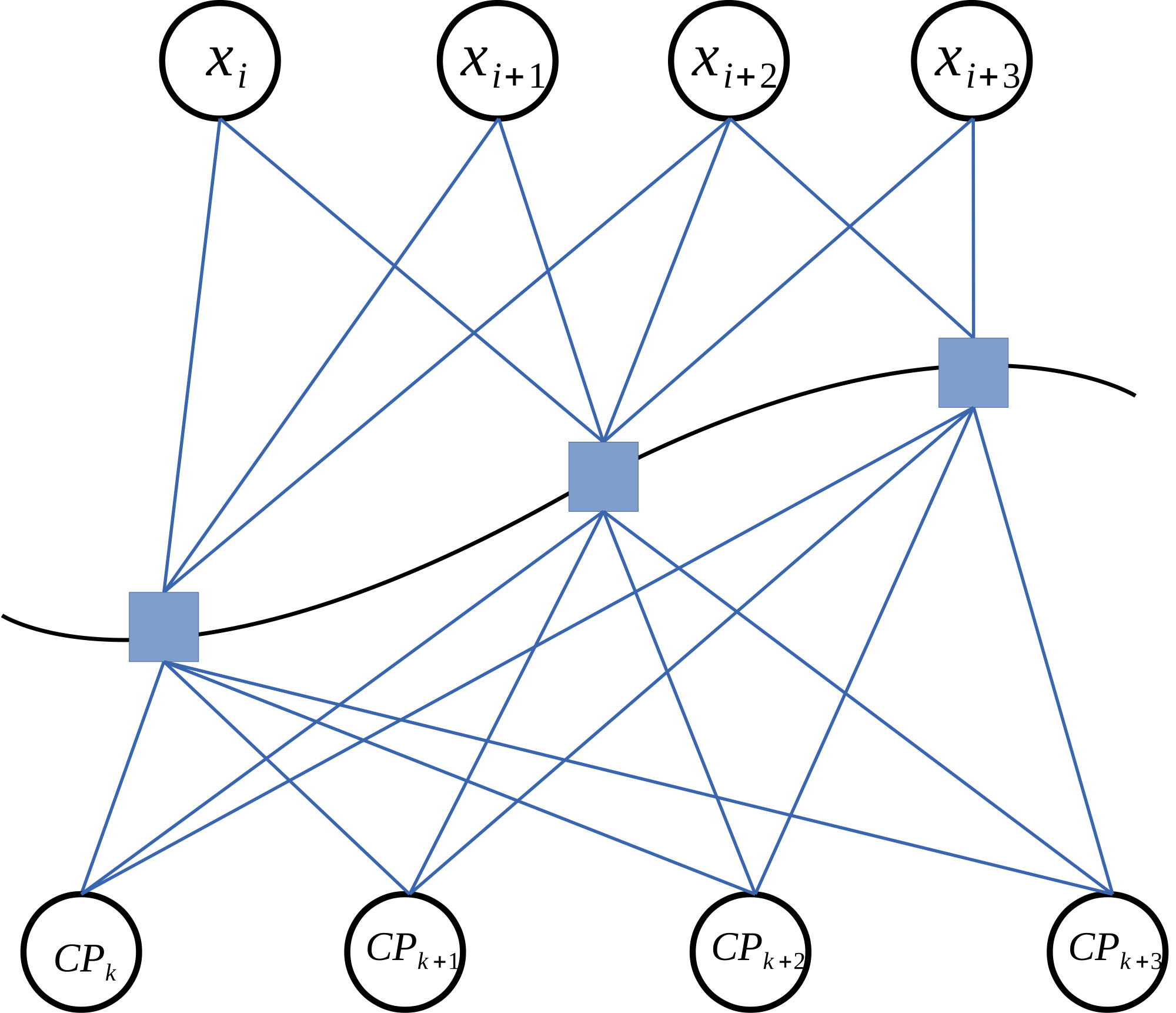}}
\vspace{-0.2cm}
\caption{Graphical models of the different methods: (a) CBA; (b) CBARt; (c) CBASpRv; (d) SSBARv; (e) FSBA.}
\vspace{-0.5cm}
\end{figure*}

\subsubsection{Conventional Bundle Adjustment (CBA)}

Conventional Bundle Adjustment (CBA) consists of minimizing reprojection errors over directly parametrized poses and landmarks. The non-linear objective is given by

\small
\begin{equation}
\min_{\tiny\begin{matrix}\{\mathbf{t}^v_{i}\}\\ \{\mathbf{R}^v_{i}\}\\ \{\mathbf{x}_k\}\end{matrix}\small}
 \underbrace{\sum_{l,k,i} \rho\left( \|\pi_{c_l}\left(\mathbf{T}_{c_lv} \left[\begin{matrix} \mathbf{R}^v_i & \mathbf{t}^v_i\\ \mathbf{0} & 1 \end{matrix}\right]^{-1} \mathbf{x}_k\right)-\mathbf{u}^{c_l}_{ki}\|^2 \right),}
_{\text{conventional bundle adjustment (CBA)}}
\label{eq:conventionalba}
\end{equation}
\normalsize
where $\mathbf{t}^v_i$ and $\mathbf{R}^v_i$ are the optimised pose parameters; $\{\mathbf{x}_k\}$ are the optimized landmarks (now in homogeneous representation); $\mathbf{u}^{c_l}_{ki}$ is the measurement of the $k$-th landmark into camera $c_l$ at time index $i$; $\rho\left(\cdot\right)$ is a loss function (e.g. Huber loss) to mitigate the influence of outliers; and $\pi_{c_l}\left(\cdot\right)$ is our camera model that transforms points from the camera frame (now in homogeneous form) into the image plane. $\mathbf{T}_{c_lv}$ still represents the extrinsic parameters that transform points from the vehicle to the camera frame. The graphical model of this problem is shown in Figure \ref{fig:CBA}. Blue nodes are re-projection error residual blocks.

\subsubsection{CBA with R-t constraints (CBARt)}

The traditional way consists of adding the R-t constraint \eqref{eq:rtconstraint} as a pairwise, soft regularization constraint to bundle adjustment (cf.~\cite{zong2017vehicle} and~\cite{li18}). We obtain

\small
\begin{align}
    \min_{\tiny\begin{matrix}\{\mathbf{t}^v_{i}\}\\ \{\mathbf{R}^v_{i}\}\\ \{\mathbf{x}_k\}\end{matrix}\small}
    &\underbrace{\sum_{l,k,i} \rho\left( \|\pi_{c_l}\left(\mathbf{T}_{c_lv} \left[\begin{matrix} \mathbf{R}^v_i & \mathbf{t}^v_i\\ \mathbf{0} & 1 \end{matrix}\right]^{-1} \mathbf{x}_k\right)-\mathbf{u}^{c_l}_{ki}\|^2 \right)}_{\text{conventional bundle adjustment (CBA)}} \\ \notag
+ & \underbrace{\sum_{\{i,j\}}w_r\| \left(\left(\mathbf{I}+\mathbf{R}^v_{ij}\right)\tiny\left[\begin{matrix}0\\1\\0\end{matrix}\right]\small\right) \times \mathbf{t}^v_{ij} \|^2}_{\text{R-t constraint}},
\end{align}
\normalsize
where $w_r$ is a scalar weight for the R-t constraints, and the latter depend on the relative rotation $\mathbf{R}^v_{ij}=\mathbf{R}_i^{vT}\mathbf{R}_j^v$ and the relative translation $\mathbf{t}^v_{ij}=\mathbf{R}^{vT}_i(\mathbf{t}^v_j-\mathbf{t}^v_i)$ between pairs of subsequent views. The objective still optimises a discrete set of poses and regularizes it against a piece-wise circular arc model. The graphical model of this problem is shown in Figure \ref{fig:CBA_Rt}. Yellow nodes indicate the R-t constraints.

\subsubsection{CBA with spline regression (CBASpRv)}

We proceed to our first utilization of a continuous time model, where we still use CBA to optimise individual poses, but interleavingly regress a 3D spline to the optimised positions $\mathbf{t}_b$ which we then use in a soft, regularising R-v constraint that replaces the R-t constraint. Denoting the alternatingly updated spline by $\mathbf{c}_1(t)$, the objective now becomes

\small
\begin{align}
& \min_{\tiny\begin{matrix}\{\mathbf{t}^v_i\}\{\mathbf{R}^v_i\} \\ \{\mathbf{x}_k\},\mathcal{P}\end{matrix}\small}
\underbrace{\sum_{l,k,i} \rho\left( \|\pi_{c_l}\left(\mathbf{T}_{c_lv}\left[\begin{matrix} \mathbf{R}^v_i & \mathbf{t}^v_i\\ \mathbf{0} & 1 \end{matrix}\right]^{-1} \mathbf{x}_k\right)-\mathbf{u}^{c_l}_{ki}\|^2 \right)}
_{\text{conventional bundle adjustment (CBA)}} \\ \notag
& + \underbrace{\sum_{i}w_s\|\mathbf{t}^v_i-\mathbf{c}_1(t_i)\|^2}_{\text{smoothness constraint}}
+ \underbrace{\sum_{i}w_c\|\mathbf{R}^v_i\tiny\left[\begin{matrix}0\\1\\0\end{matrix}\right]\small-\eta(\mathbf{c}'_1(t_i))\|^2,}_{\text{R-v constraint}}
\end{align}
\normalsize
where $\mathcal{P}$ is the set of control points of $\mathbf{c}_1(t)$; $t_i$ is the timestamp for $i$th vehicle pose; $w_s, w_c$ are scalar weights;
$\eta(\mathbf{a})=\frac{\mathbf{a}}{\|\mathbf{a}\|}$;
and $\mathbf{c}'(t)$ denotes the first-order derivative of $\mathbf{c}(t)$.
Note that the latter is readily given as a spline that sums over products between control points and the fixed, first-order derivatives of the basis functions. The graphical model of this problem is shown in Figure \ref{fig:CBA_Rv}. The green nodes are the combined smoothness and R-v constraints. $\mathit{CP}$ are the control points.

\subsubsection{Soft spline bundle adjustment (SSBARv)}\label{SSBA}

In our next formulation, the spline is used directly to represent the pose. Only the kinematic R-v constraint remains as side-constraint. 
We use the 7D spline $\mathbf{c}_2(t)=\left[\small\begin{matrix}\mathbf{c}^{\mathbf{t}}_2(t) \\ \mathbf{c}^{\mathbf{q}}_2(t)\end{matrix}\normalsize\right]$ which represents the position in its first three entries and the quaternion orientation in its remaining four, see~\cite{kang1999cubic} for details about unit quaternion B-splines. We obtain

\small
\begin{align}
\min_{\{\mathbf{x}_k\}, \mathcal{P}}& \underbrace{\sum_{l,k,i} \rho\left( \|\pi_{c_l}\left(\mathbf{T}_{c_lv}
\left[\begin{matrix} \mathbf{R}(\mathbf{c}_2^{\mathbf{q}}(t_i)) & \mathbf{c}_2^{\mathbf{t}}(t_i)\\ \mathbf{0} & 1 \end{matrix}\right]^{-1}
\mathbf{x}_k\right)-\mathbf{u}^{c_l}_{ki}\|^2 \right)}_{\text{7D spline bundle adjustment(SSBA)}} \\ \notag
& + \underbrace{\sum_{j}w_c\| \mathbf{R}(\mathbf{c}_{2}^{\mathbf{q}}(t_i))\tiny\left[\begin{matrix}0\\1\\0\end{matrix}\right]\small-\eta(\mathbf{c}^{\mathbf{t}'}_2(t_i)) \|^2,}_{\text{R-v constraint}}
\end{align}
\normalsize
where $\mathcal{P}$ is the set of control points of $\mathbf{c}_2(t)$, and $\mathbf{R}(\cdot)$ is defined as a function that takes a quaternion and returns the corresponding rotation matrix. The graphical model of this problem is shown in Figure \ref{fig:SSBA_Rv}. The red nodes are the R-v constraints.

\subsubsection{Hard spline bundle adjustment (FSBA)}

Our final objective consists of directly using the derivative of the continuous position to express part of the vehicle orientation (i.e. the heading). As a result, we employ the four-dimensional spline $\mathbf{c}_3(t) = \small\left[ \begin{matrix} \mathbf{c}_3^{\mathbf{t}}(t) \\ \alpha(t) \end{matrix} \right]\normalsize$, where the first three entries denote the vehicle position as before, its derivative will be used to obtain the heading, and $\alpha(t)$ is a 1D spline that models the remaining roll angle about the heading direction. Let's denote the vehicle orientation at time $t$ as $\mathbf{U}(\mathbf{c}_3(t))$. It is defined as

\small
\begin{equation}
  \mathbf{U}(\mathbf{c}_3(t))=\mathbf{Q}(\mathbf{c}_3^{\mathbf{t}}(t))\left[\begin{matrix}  \cos\alpha(t) & 0 & \sin\alpha(t) \\ 0 & 1 & 0 \\ -\sin\alpha(t) & 0 & \cos\alpha(t) \end{matrix}\right],
\end{equation}
\normalsize
where $\mathbf{Q}(\mathbf{c}_3^{\mathbf{t}}(t))$ is the roll-free base orientation and defined as
\small
\begin{equation}
  \mathbf{Q}(\mathbf{c}_3^{\mathbf{t}}(t)) = \left[\begin{matrix}
      \eta\left(\eta(\mathbf{c}_3^{\mathbf{t}'}(t)) \times \tiny\left[\begin{matrix}0\\0\\1\end{matrix}\right]\small\right)^T \\
      \eta(\mathbf{c}_3^{\mathbf{t}'}(t))^T \\
      \left(\eta\left(\eta(\mathbf{c}_3^{\mathbf{t}'}(t)) \times \tiny\left[\begin{matrix}0\\0\\1\end{matrix}\right]\small\right) \times \eta(\mathbf{c}_3^{\mathbf{t}'}(t))\right)^T
      \end{matrix}\right]^T.
\end{equation}
\normalsize

The base orientation is defined such that the heading---the second column of $\mathbf{Q}(\mathbf{c}^\mathbf{t}_3(t))$---is aligned with the first-order differential of the vehicle trajectory, and the side-ways direction---the first column of $\mathbf{Q}(\mathbf{c}^\mathbf{t}_3(t))$---is orthogonal to the vertical direction $[0\text{ }0\text{ }1]^T$. The objective becomes

\small
\begin{equation}
  \min_{\{\mathbf{x}_k\}, \mathcal{P}}
  \underbrace{\sum_{l,k,i} \rho\left( \|\pi_{c_l}\left(\mathbf{T}_{c_lv}
  \left[\begin{matrix} \mathbf{U}(\mathbf{c}_3(t_i)) & \mathbf{c}_3^{\mathbf{t}}(t_i) \\ \mathbf{0} & 1\end{matrix}\right]^{-1}
  \mathbf{x}_k\right)-\mathbf{u}^{c_l}_{ki}\|^2 \right)}
  _{\text{4D spline bundle adjustment(FSBA)}}
  \label{eq:FSBA}
\end{equation}
\normalsize
where $\mathcal{P}$ now denotes the set of control points of $\mathbf{c}_3(t)$. The graphical model of this problem is shown in Figure \ref{fig:FSBA}.


\section{Full surround-view SLAM system}
\label{sec:multivo}

Our final contribution consists of a surround-view simultaneous tracking and mapping framework built around two of the core modules described in prior sections: the motion initialization method from Section~\ref{sec:vo} as well as the spline-based back-end optimization from Section~\ref{sec:optimization}. The framework employs the ORB feature~\cite{rublee2011orb} and its tracking and mapping modules are strongly inspired by the ORB-SLAM framework~\cite{murORB2}. The system contains two working threads, that is, a front-end thread running visual odometry, and a back-end optimization thread running windowed bundle adjustment. Note however that no loop-closure module is added. Rather, we install an optional but careful inclusion of low-grade GPS signals in order to prevent long-term drift accumulation. Owing to the employed core modules, the framework achieves highly reliable initialization of relative displacements in the front-end and accurate, instanteous vehicle kinematics-aware optimization in the back-end.

In the front-end visual odometry thread, each camera from the surround-view system is first handled independently to extract ORB features in each incoming frame. Note that both the original version from OpenCV as well the modified version from ORB-SLAM (aiming for more homogeneous feature distributions) can be chosen\footnote{The framework simply uses adapters to OpenCV~\cite{opencv_library} feature detection and description methods, hence any feature contained in the OpenCV library combined with a suitable matcher can be used}. Once features are extracted, we then perform brute-force temporal intra-camera matching. No inter-camera stereo matching is performed as the overlap area between neighbouring fields of view is deemed absent or at least too small and too distorted for reliable feature extraction and matching. Instead, metric scale is recovered and maintained via the optional inclusion of GPS readings (details below).

\begin{figure}[t]
\begin{center}
\vspace{1 ex}
\includegraphics[width=\linewidth]{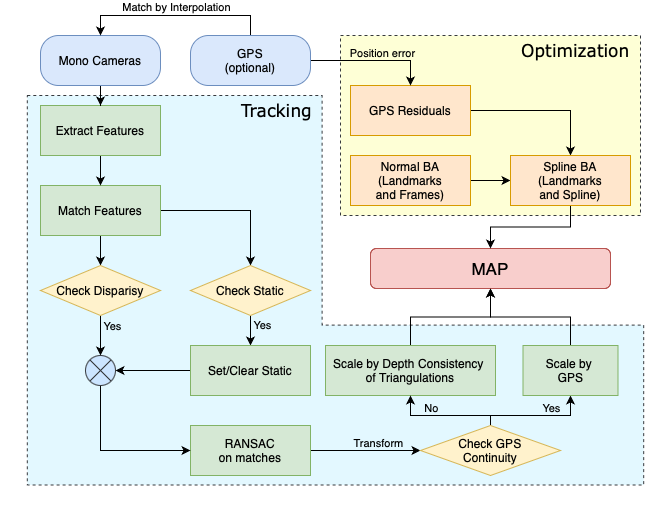}
\end{center}
\caption{Surround-View SLAM system overview, showing all the steps performed by the tracking and back-end optimization threads.}
\label{fig:multivo}
\end{figure}

Not every frame is handed to the back-end optimizer. Instead, the front-end employs a simple keyframe generation mechanism that verifies if the median of the frame-to-frame disparities has surpassed a certain threshold. The method from Section~\ref{sec:vo} is then used to compute relative transformations within RANSAC thereby discovering all inlier correspondences. Subsequently, two cases need to be distinguished. If a correspondence is new (i.e. no prior existing reference to a 3D landmark), a new 3D landmark is triangulated. Conversely, if a reference exists, the list of measurements for that landmark is simply extended by the new measurement from the most recently added keyframe. Note furthermore that existing landmarks are used to consistently rescale the magnitude of the translational displacement between the new and the previous keyframe. Given multiple such landmarks may exist, we again perform such action in a robust manner by taking the median over all rescaling factors.

The mapping thread triggers a new round of optimization whenever enough new keyframes have been added. The optimization starts with a standard bundle adjustment implementation to refine keyframe poses and landmarks by minimizing the reprojection errors~\eqref{eq:conventionalba}. This pre-optimization can provide a reliable initial guess for the subsequent spline optimization. The latter employs the superior, hardly constrained FSBA variant (cf.~\eqref{eq:FSBA}) to ensure an accurate, smooth, kinematics-aware estimation result. Given no inertial readings with gravity measurements is present, we furthermore add a regularization term on the rolling angle to limit drift along the fourth dimension. To conclude, GPS data-terms are added as an optional position residual to mitigate long-term positional drift.

The inclusion of kinematically satisfied trajectory models requires a careful temporal knot placement strategy. This is particularly the case given that long trajectories may contain passages with fewer turns, while other segments may require more knots and control points to allow for higher non-linearities. In our work, we consider the automatic knot placement algorithm introduced in (9.8) in~\cite{piegl2012nurbs}, which considers a simple averaging strategy that depends on the local temporal density of the input samples. The latter is relatively homogeneous because as soon as the vehicle surpasses a certain velocity, every incoming set of surround-view images will trigger a new keyframe. The advantage of this strategy is that the corresponding control points will be implicitly more spaced on straights where vehicle speed tends to be more elevated, and less spaced when the vehicle enters a turn. The down-side of the strategy however is given by the fact that in urban environments, traffic lights and pedestrian crossings cause the speed of the vehicle to not always be correlated with the complexity of the road layout.

An extreme case is attained when the car comes to a complete stop. In such periods, automatic knot placement may trigger additional knots despite a longer time interval of immobility. This in turn will lead to a local accumulation of control points, and severe artefacts in the final optimization result. In order to circumvent this problem, a special module detecting static intervals is added to the front end. The module simply considers the frame-to-frame disparity of all feature matches, and evaluates what total fraction has a negligible disparity by comparing against a very small threshold. If that fraction is large enough, movement with respect to the background is considered absent. Intervals of inactivity are logged, and the internal time-line used for curve-parametrization is a continous concatenation of the active intervals from the original time line, only.

As mentioned before, low-grade GPS readings are optionally integrated for more robust localization. Care needs to be taken in order to avoid the use of sudden outlier GPS readings emanating from a change in the number of received satellite signals or multi-path effects~\cite{van1992multipath}. Jumps are successfully detected by the continuity of GPS readings. A GPS data term is added only if it remains within expected GPS error bounds.

The proposed configuration ensures reliable, globally consistent mapping and localization even in the absence of loop closure or supplementary sensors, which renders it easily deployable on existing platforms. The framework does not depend on tedious IMU calibration or the inter-connection with vehicle inherent sensors such as wheel odometers.

\section{Experimental results}
\label{sec:results}

The experimental results section divides into four parts. The first three sections focus on ablation studies of the three core modules presented in this work, and analyze their performance as a function of additive noise. We furthermore underline our relative pose solver's ability to deal with deviations from the pure Ackermann model, and analyze the robustness of our back-end model with respect to varying connectivity in the bundle adjustment graph. To conclude, we present large-scale results of our full surround-view camera SLAM system obtained on the public benchmark \textit{Oxford RobotCar}~\cite{RobotCarDatasetIJRR}.

\subsection{Ablation studies of online calibration}

The experimental evaluation of the online calibration contains two parts. We start with simulations in which we test the method's resilience against noise. We also apply our method to calibrate the cameras used to collect the publicly available KITTI stereo benchmark \cite{geiger2013vision}. In both experiments, we use the relative pose solver presented in \cite{stewenius06} embedded into a Ransac \cite{fischler81} scheme to robustly recover relative camera-to-camera displacements. We notably use the implementation contained in OpenGV \cite{kneip14opengv}.

\subsubsection{Verification in simulation}
We generate simulation experiments by assuming that there are two cameras mounted on a vehicle moving on a plane. They are facing to the left and right, have horizontal principal axes', a baseline of 0.1$m$, image sizes of 500$\times$500 pixels, and focal lengths of 250 pixels. We generate sequences of 60 virtual camera frames with 0.3$m$ frame-to-frame spacing. The vehicle is moving straight during the first 20 frames, and taking turns to either side during the remaining part of the datasets. Image measurements are produced by generating random 3D points and vertical lines and projecting them into each frame. We conclude with adding uniform noise on the image observations. Besides resilience against noise, the sequences test the algorithms ability to automatically distinguish between straight and curved trajectory segments.

\begin{figure}[t]
\begin{center}
\vspace{1 ex}
\includegraphics[width=\linewidth]{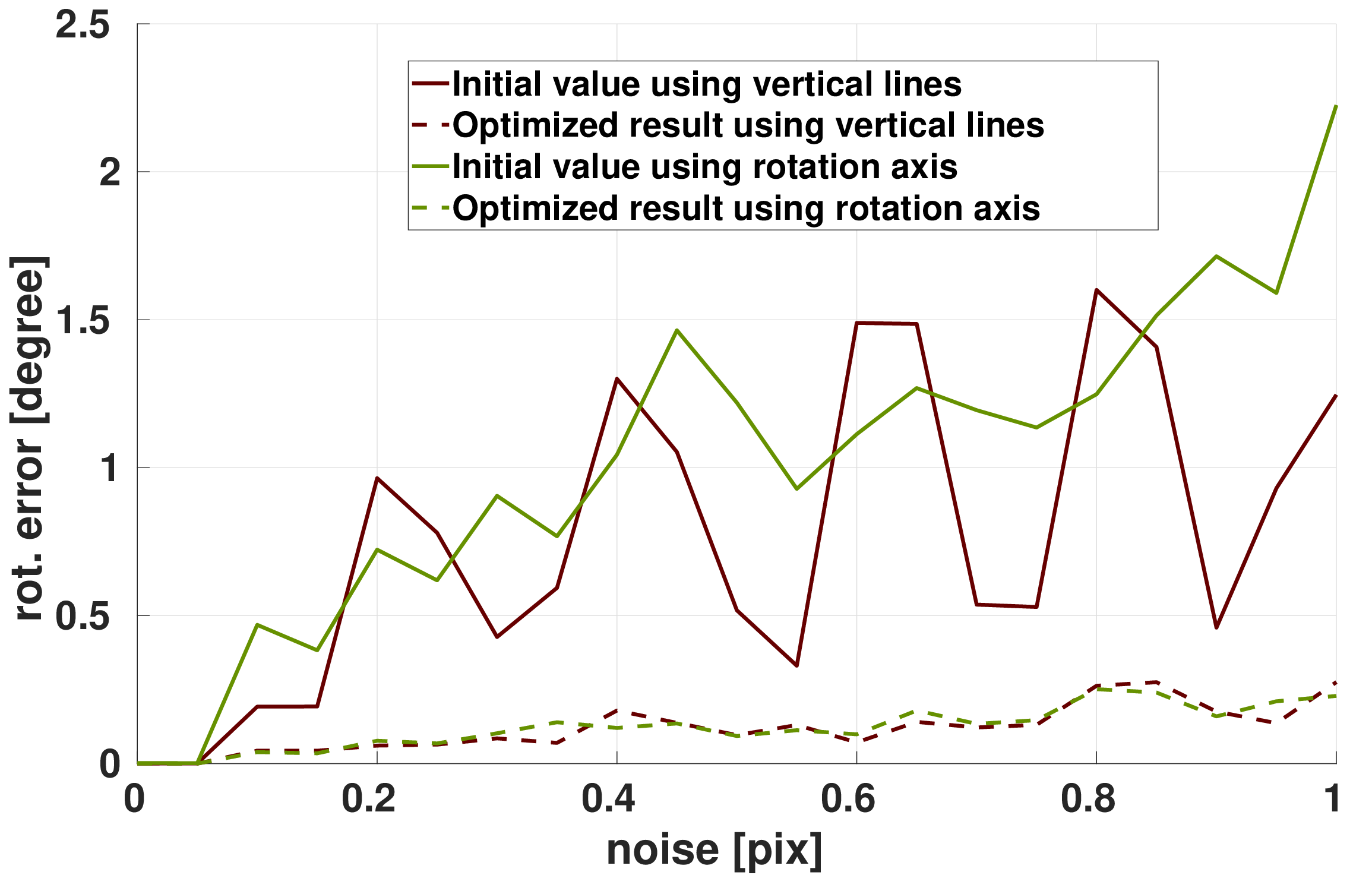}
\end{center}
\caption{Accuracy of the proposed calibration method as a function of image noise. The plot indicates both the quality of the initialization from a minimal number of views (solid lines), as well the final error after multi-frame optimisation (dashed lines). Both rotation axes (green) and vertical lines (red) can be used.}
\label{fig:simulation}
\end{figure}

Our result is indicated in Figure \ref{fig:simulation}. Green lines indicate the result obtained by using measurements of the rotation axis, while red lines indicate the result obtained by using measurements of vertical lines. Solid lines indicate initial errors, while the dashed lines indicate the final result after optimisation. Initial results are simply obtained by considering the translation vector and the vertical vanishing point obtained from a single pair of views. As can be observed, despite larger variations in the quality of the initialization values, the method consistently proves a good ability to average out noise over the  entire sequence. Furthermore, it can be concluded that under similar noise properties, there is no noticeable difference between using either the rotation axis or vertical line measurements.

\subsubsection{Tests on KITTI}
In order to further test the ability of exploiting vertical line measurements on natural data, we apply our method to the stereo sequences provided by the KITTI benchmark datasets \cite{geiger2013vision}, and disable the rotation axis constraint ($\lambda_2=0$). A qualitative evaluation is obtained by ignoring the overlap in the cameras' fields of view, calibrating the camera-to-body rotation for each camera individually, and concatenating them to obtain a measure of the direct camera-to-camera transformation. The latter is compared against the ground-truth value provided on the web-page. We evaluate our method over multiple sequences, each time choosing 10 different sub-sequences of 70 frames. Average errors for each sequence are indicated in Table \ref{table:KITTI-results}. As can be observed, the approach consistently leads to errors below 2$^{\circ}$. Note that the deviation from the values provided on the webpage is somewhat consistent suggesting that part of the remaining discrepancy may indeed originate from inconsistent outcomes of different calibration methods rather than an error with respect to (unavailable) ground truth.
\begin{table}[b]
\caption{Calibration results obtained on KITTI sequences. Errors are indicated in degrees, and indicate the deviation from ground-truth of deduced camera-to-camera rotations.}
\begin{center}
\begin{tabular}{|c|c|c|}
\hline
Sequence & Initial value & optimised result\\
\hline
0046 & $2.5341^{\circ}$ &  $1.7025^{\circ}$\\
\hline
0064 &   $1.8620^{\circ}$ &   $1.4345^{\circ}$\\
 \hline
0104(part 1) &   $2.8485^{\circ}$&    $1.5084^{\circ}$\\
 \hline
0104(part 2) &  $2.1060^{\circ}$   & $1.4984^{\circ}$\\
\hline
\end{tabular}
\end{center}
\label{table:KITTI-results}
\end{table}

\subsection{Ablation studies of motion initialization framework}

The motion initialization framework is again tested on both synthetic and real data. Our solver depends on planar motion and is designed for non-overlapping multi-camera systems. Our experiments therefore focus on a comparison against previous relative pose solvers for generalized cameras, which are a 2-point Ransac algorithm relying on a non-holonomic motion assumption, the linearized 17-point algorithm, and a generalized eigenvalue minimization algorithm. In order to demonstrate the benefit of using multiple cameras pointing into different directions, we also include a comparison against centralized, single-camera algorithms such as the traditional 8-point algorithm and 1-point Ransac, the latter again relying on a non-holonomic motion assumption. We execute different comparative simulation experiments to evaluate accuracy and noise resilience.

\subsubsection{Outline of the experiments}

We assume the surround-view camera system has four co-planar cameras pointing into all directions. In each experiment iteration, we fix the frame of the first viewpoint to coincide with the world frame. We then add 6 further views by adopting a linearly changing rotational velocity, therefore generating realistic, non-circular motion trajectories. We finally calculate the relative displacement between the first and the last viewpoint. We generate random correspondences for each camera by defining random 3D landmarks located within the field of view of each camera in the first viewpoint. We also added noise and outliers into the measurements, and finally analyze the performance of our method under different conditions, which are a dynamic rotational velocity, purely translational motion, and a varying field-of-view. We execute 1000 random constellations for each experiment, and---due to scale observability issues---report primarily on the accuracy of the relative rotation estimation.

\subsubsection{Comparison against minimal solvers}
\begin{figure*}[t]
  \centering
  \subfigure[]{
    \label{Simulation:dev_ack}
    \includegraphics[width=0.245\textwidth]{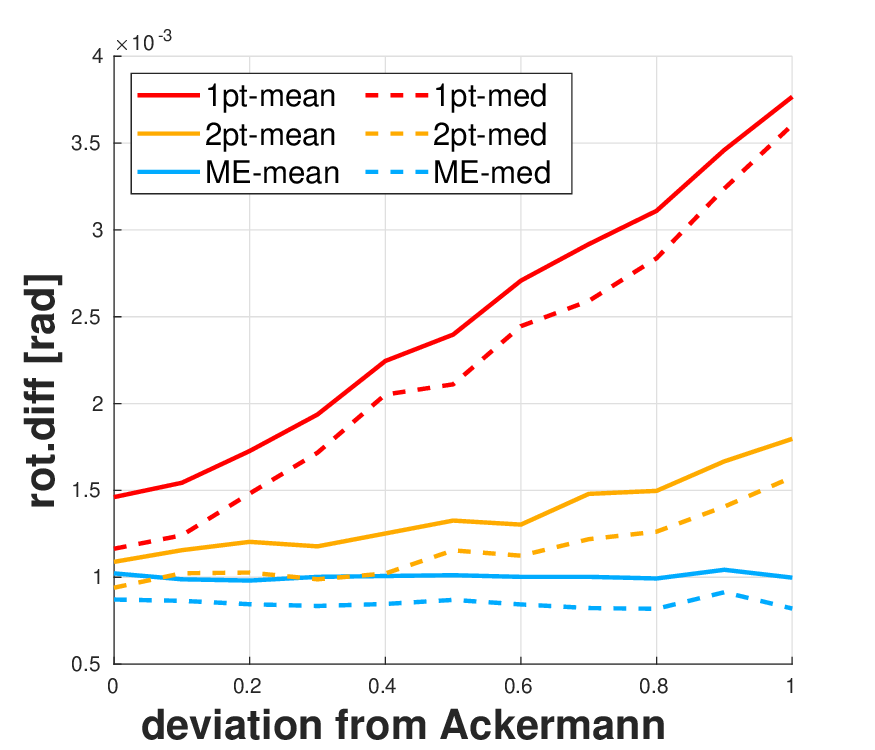}}
    \hspace{-2ex}
  \subfigure[]{
    \label{Simulation:noise}
    \includegraphics[width=0.245\textwidth]{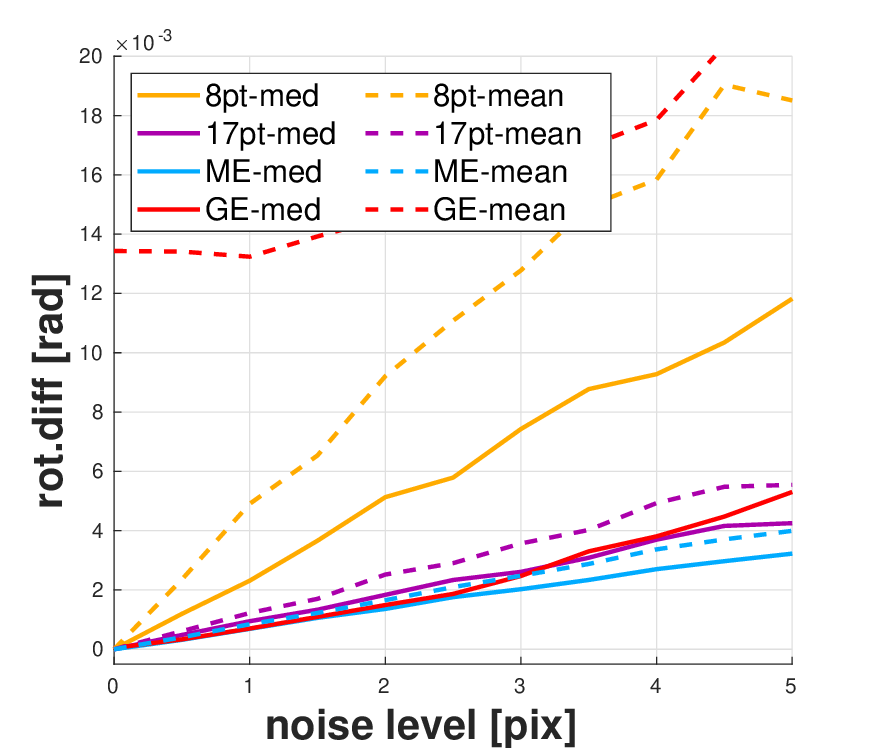}}
    \hspace{-2ex}
  \subfigure[]{
    \label{Simulation:straight}
    \includegraphics[width=0.245\textwidth]{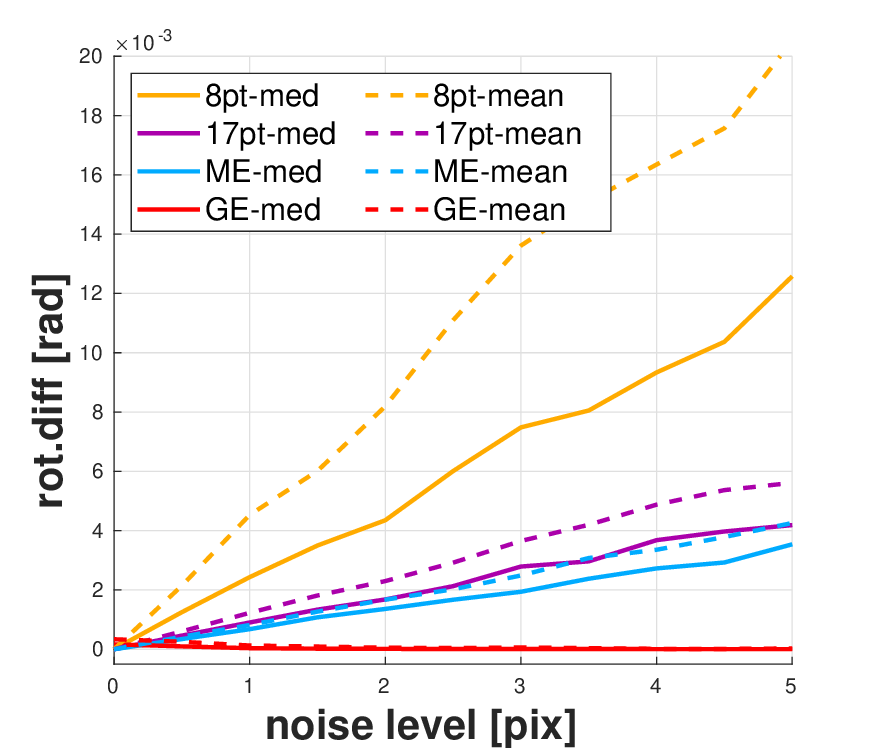}}
    \hspace{-2ex}
  \subfigure[]{
    \label{Simulation:fov}
    \includegraphics[width=0.245\textwidth]{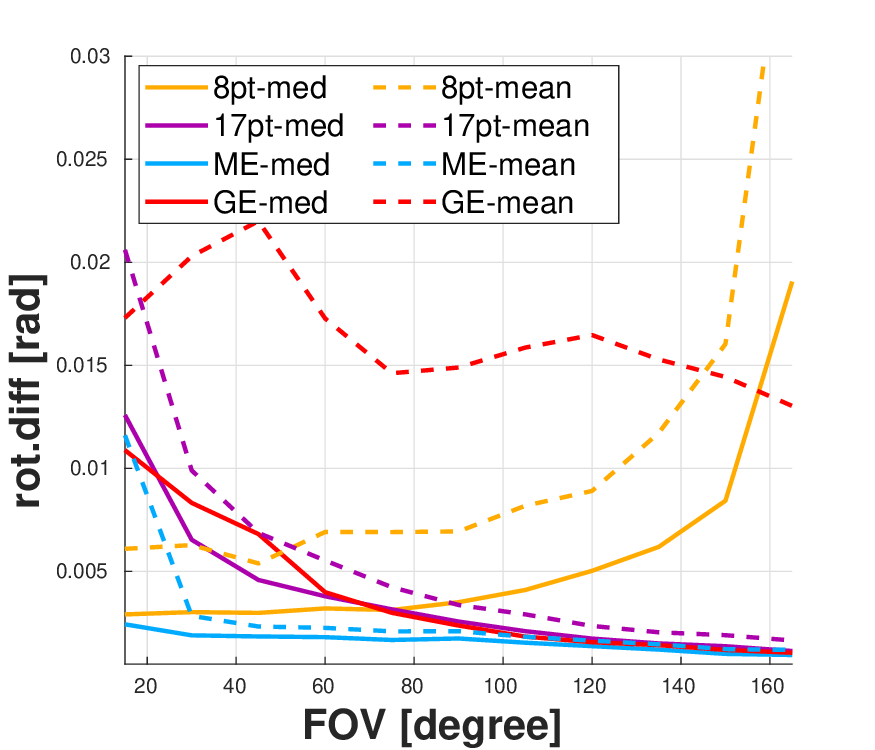}}
  \caption{Comparison between our proposed method \textbf{ME} and the \textbf{1pt}, \textbf{2pt}, \textbf{8pt}, \textbf{17pt}, \textbf{GE} method for different perturbation factors. Each value is averaged over 1000 random experiments. Details are provided in the text.}
  \vspace{-0.5cm}
  \label{Simulation}
\end{figure*}

We compare our method (\textbf{ME}) against state-of-the-art minimal solvers for planar motion, which is the 2-point Ransac algorithm by \cite{lee13} (\textbf{2-pt}) and the 1-point Ransac algorithm proposed by \cite{scaramuzza09} (\textbf{1-pt}). These algorithms adopt the Ackermann motion model, and thus make the assumption that the motion is non-holonomic with a fixed centre of rotation (ICR) for the entire relative displacement. As a result, they have a reduced number of required correspondences. We generate 20 points in each camera for all algorithms, and use the minimal number of points for each method to solve the problem hypotheses (1 point for \textbf{1-pt}, 2 points for \textbf{2-pt} and 3 points per camera in our algorithm). No outliers are added to the data. We repeat the experiment for changing deviations from pure Ackermann motion defined by the linearly changing per-frame rotation change $\omega = 0.04k \cdot i + \omega_{0}$, where $i = 1,\ldots,6$, $\omega_{0} = 0.2^\circ$, and $k$ is varied from 0 to 10. The ICR is extrapolated by assuming the constant forward velocity $v=0.1$m per second. The results are indicated in Figure \ref{Simulation:dev_ack}. As expected, our model outperforms as the deviation from non-holonomic motion is increasing.

\subsubsection{Comparison against non-minimal solvers}
We compare our method against alternative non-minimal, generalized solvers (\textbf{17-pt} \cite{li08} and \textbf{GE} \cite{kneip14}) as well as a central method (\textbf{8-pt} \cite{hartley97}) commonly applied in vehicle motion estimation with a forward-facing camera. We use 5 points in each camera for all generalized algorithms and 8 points in the forward camera for \textbf{8-pt}. We run only the solvers and do not add nonlinear refinement. We conduct three types of experiments:

\begin{itemize}
\item \textit{Significant rotation}: The field-of-view of each camera is fixed to 120$^\circ$, and the noise level is varied between 0 and 5 pixels. The results are indicated in Figure \ref{Simulation:noise}. As can be observed, all generalized solvers out-perform the centralized method, and \textbf{ME} performs better than \textbf{17-pt} in terms of both the mean and median error. As stated in \cite{kneip14}, \textbf{GE} has been designed for omnidirectional cameras and occasionally converges into wrong local minima, thus leading to an increased mean error with repect to \textbf{ME}.
\item \textit{Pure translations}: We repeat the same experiment but simply force the motion to be purely translational. The result is illustrated in Figure \ref{Simulation:straight}. As expected, \textbf{17-pt} and \textbf{ME} maintain a higher level of accuracy than \textbf{8-pt}. As furthermore explained in \cite{kneip14}, \textbf{GE} is affected by a constant zero energy for identity rotation. While this leads to perfect performance in this experiment, it is to be interpreted as a weakness. \textbf{GE} is unable to distinguish small from zero rotation angles.
\item \textit{Variation of the field of view}: We vary the field-of-view from 15$^\circ$ to 165$^\circ$. As shown in Figure \ref{Simulation:fov}, the centralized method \textbf{8-pt} applied in the forward facing camera performs better for very small fields-of-view. As explained in Section \ref{sec:theory}, side-ways looking cameras are affected by the rotation-translation ambiguity. The effect worsens for a decreasing field-of-view, and potentially affects all generalized camera solvers. However, as soon as the field-of-view is sufficiently large (75$^\circ$ for our settings), generalized solvers start to outperform. \textbf{ME} furthermore clearly outperforms other methods and beats \textbf{8-pt} for any FoV larger than 30$^\circ$.
\end{itemize}

\subsubsection{Behavior of object-space error based refinement}
\label{sec:objVSBA}

\begin{figure}[t]
    \centering
    \includegraphics[width=0.85\linewidth]{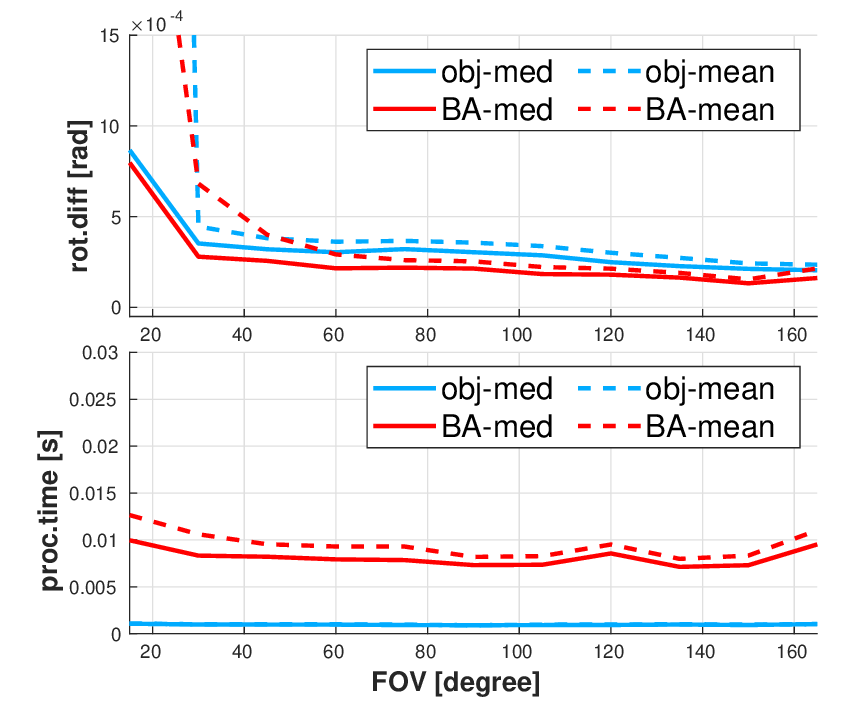}
    \caption{Accuracy of the different geometric optimization method and average execution time.}
    \vspace{-0.5cm}
    \label{fig:objVSBA}
\end{figure}

Figure \ref{fig:objVSBA} shows the comparison between our proposed object-space error minimizer and standard two-view bundle adjustment. Both depend on a sufficiently good initialization. However, as stated in Section \ref{sec:obj-space}, standard two-view bundle adjustment reduces reprojection errors over rotation, translation, and structure parameters, while the proposed joint eigenvalue minimization based object-space error reduction involves only the rotational degrees of freedom (which---in the case of planar motion---is only a single degree of freedom). As indicated in Figure \ref{fig:objVSBA}, object-space error minimization shows comparable performance than 2-view bundle adjustment for varying fields-of-view. However, owing to its univariate nature, the proposed objective is minimized 8 times faster.

\subsection{Comparison of the different back-end optimization variants}
\label{subsec:opt_variants}
Our final ablation study investigates the differences between the various frame and spline-based back-end optimization variants proposed in this work. We mainly use the KITTI benchmark datasets \cite{Geiger2012CVPR}, which are fully calibrated and contain images captured by a forward-looking camera mounted on a passenger vehicle driving through different environments. The datasets contain signals from high-end GPS/IMU sensors, which allow us to compare our results against ground truth. We use several different sequences that provide a mix of motion characteristics reaching from significant turns and height variations to simple forward motion. We also add a few carefully chosen, synthetic sequences to provide further analysis. We test all aforementioned methods plus CBASp and SSBA, which are similar to CBASpRv and SSBARv, respectively, but do not contain the kinematic R-v constraint. All our optimization variants make use of the Ceres \cite{ceres-solver} optimization toolbox with automatic differentiation.

The main purpose of this study is to demonstrate the advantage of enforcing kinematic constraints in handling degrading visual conditions. Besides the commonly analysed influence of noise on the image measurements $\mathbf{m}_{ij}$, we additionally analyse the influence of the connectivity of the graph by varying the number of landmarks and the number of observations. For each analysis and noise or connectivity setting, we calculate the mean and standard deviation of the sliding pair-wise Relative Pose Error (RPE) with respect to ground truth, which individually analyses rotation and translation errors. The rotation error is calculated using (2.15) in \cite{ma2012invitation}. For the translation error, our evaluation differs from the one in \cite{sturm12iros} in that we ignore the scale of the relative translations which is unobservable in a monocular setting.

\subsubsection{Results on synthetic data}

We start by defining realistic trajectories, which we take straight from the ground truth trajectories from KITTI sequences \cite{Geiger2012CVPR}. We also adopt the intrinsic and extrinsic parameters $f_p(\cdot)$ and $\mathbf{T}_{sb}$ from the KITTI platform, respectively. However, rather than using the original image information, we generate synthetic correspondences by defining uniformly distributed random image points in each view. The number of points denotes the local connectivity. We define random depths for these points by sampling from a uniform distribution between 6 and 30 meters. The corresponding world points (landmarks) are finally projected into all nearby views to generate all possible correspondences in the graph. Note that the number of observations per landmark is however capped by the global connectivity setting. We also perform a boundary check to make sure that reprojected points are visible in the virtual views. Finally, we add zero-mean normally distributed noise to the observations.

\begin{figure*}[ht]
\centering
\vspace{0.1cm}
\includegraphics[width=0.65\linewidth]{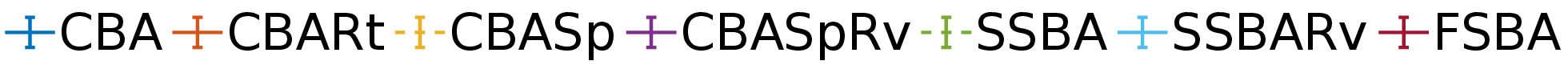}\\
\subfigure[]{\label{fig:sim-05-noise-R}\includegraphics[width=0.16\linewidth]{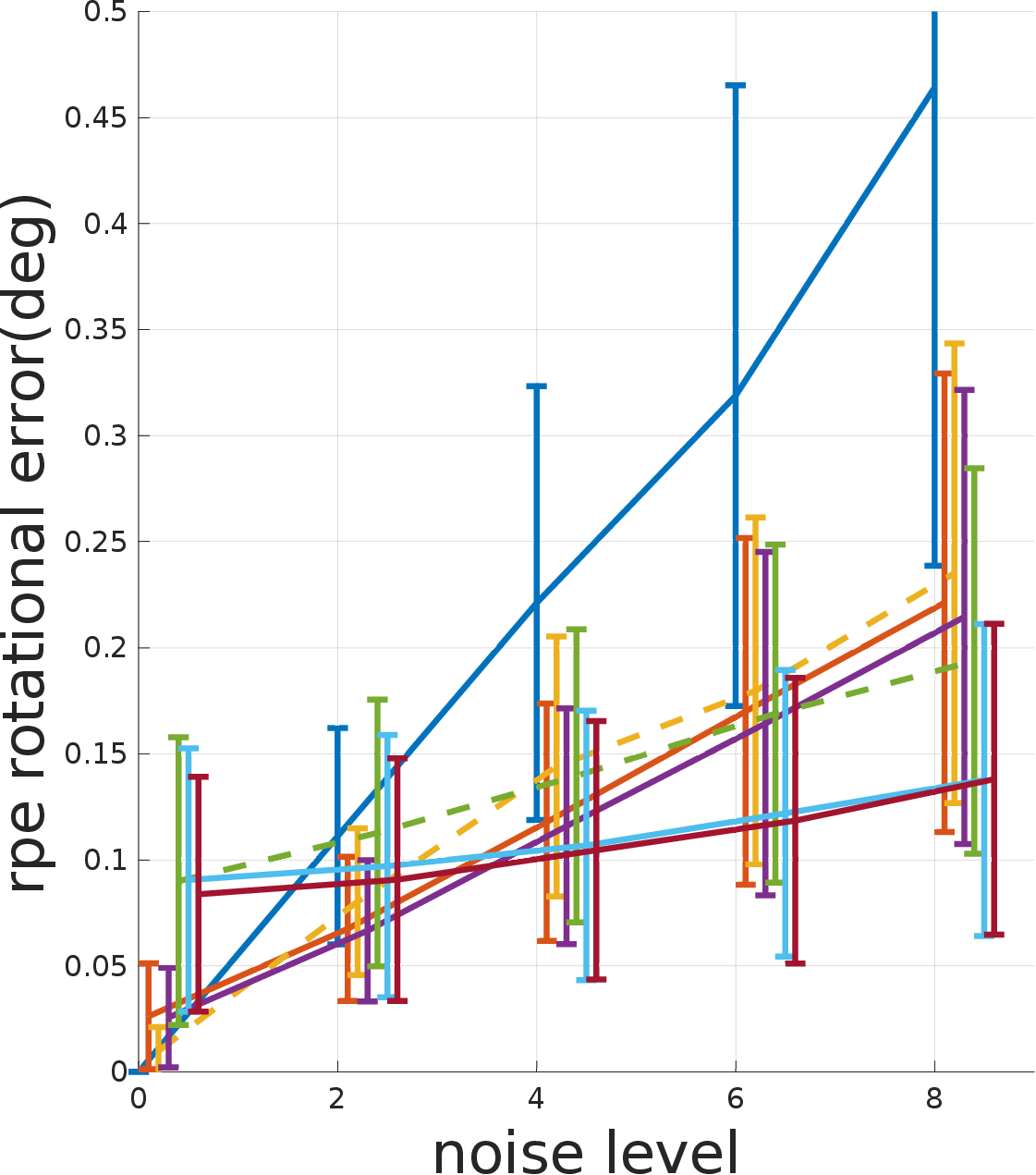}}
\subfigure[]{\label{fig:sim-05-noise-t}\includegraphics[width=0.16\linewidth]{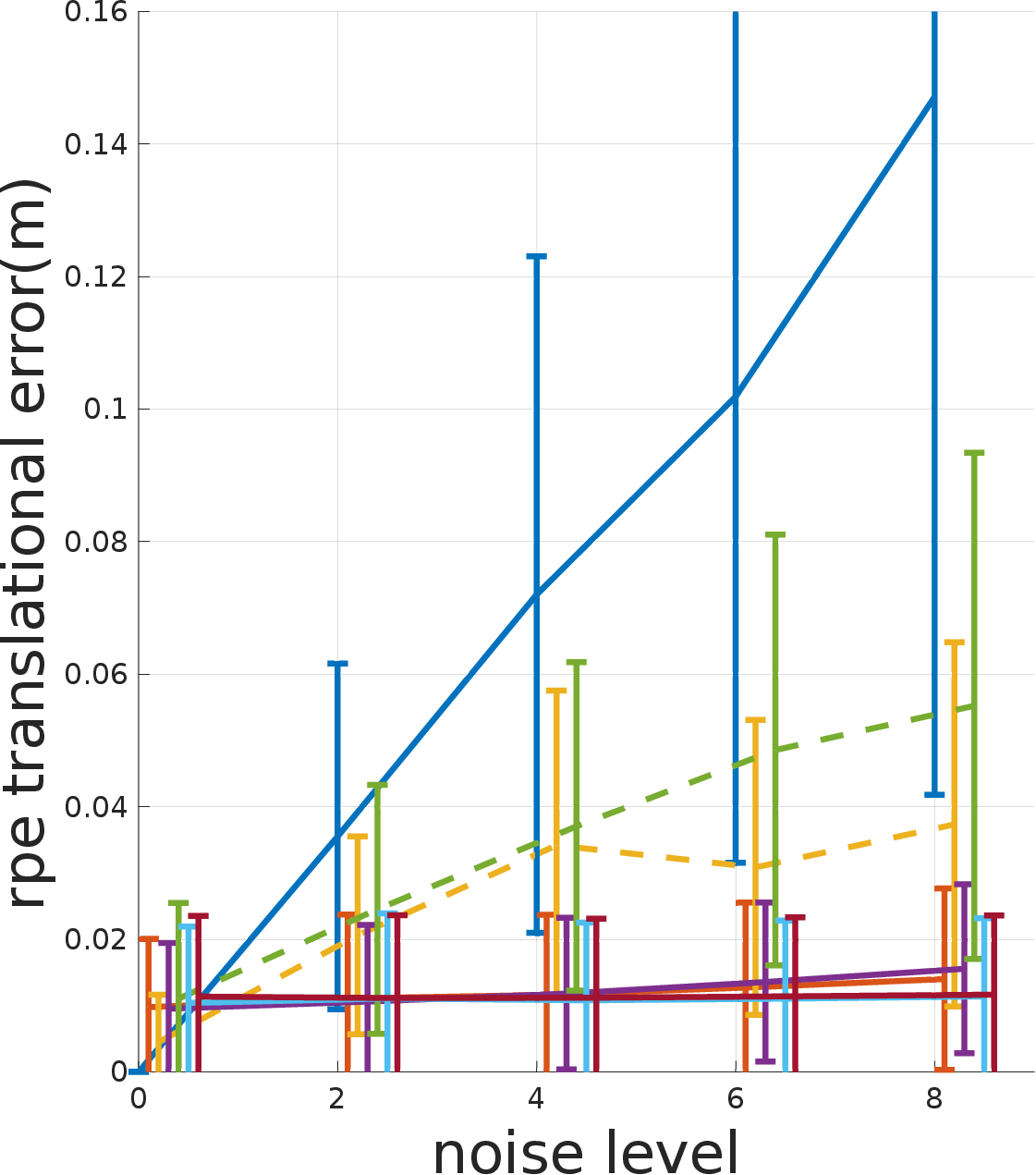}}
\subfigure[]{\label{fig:sim-05-globalC-R}\includegraphics[width=0.16\linewidth]{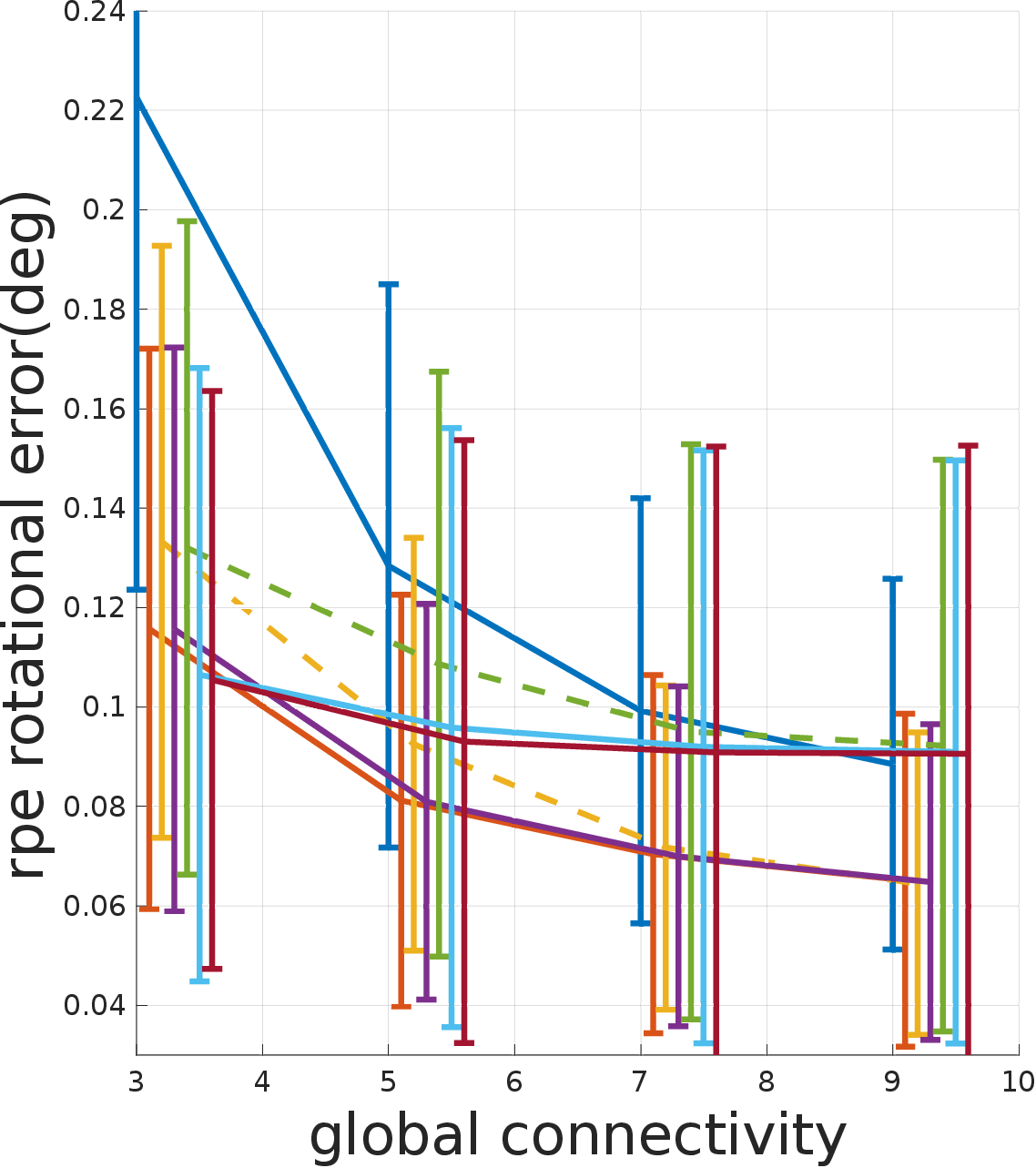}}
\subfigure[]{\label{fig:sim-05-globalC-t}\includegraphics[width=0.16\linewidth]{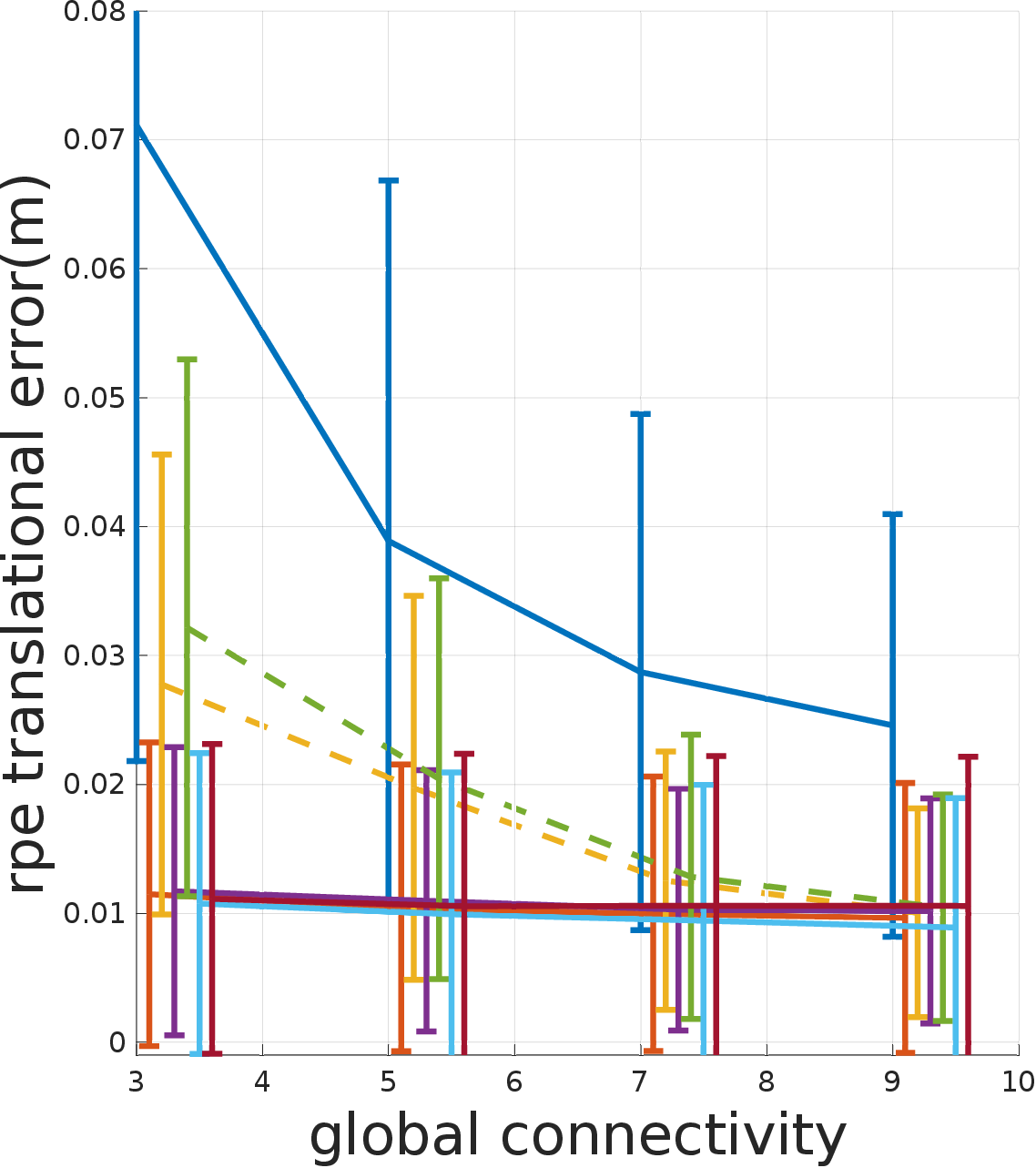}}
\subfigure[]{\label{fig:sim-05-localC-R}\includegraphics[width=0.16\linewidth]{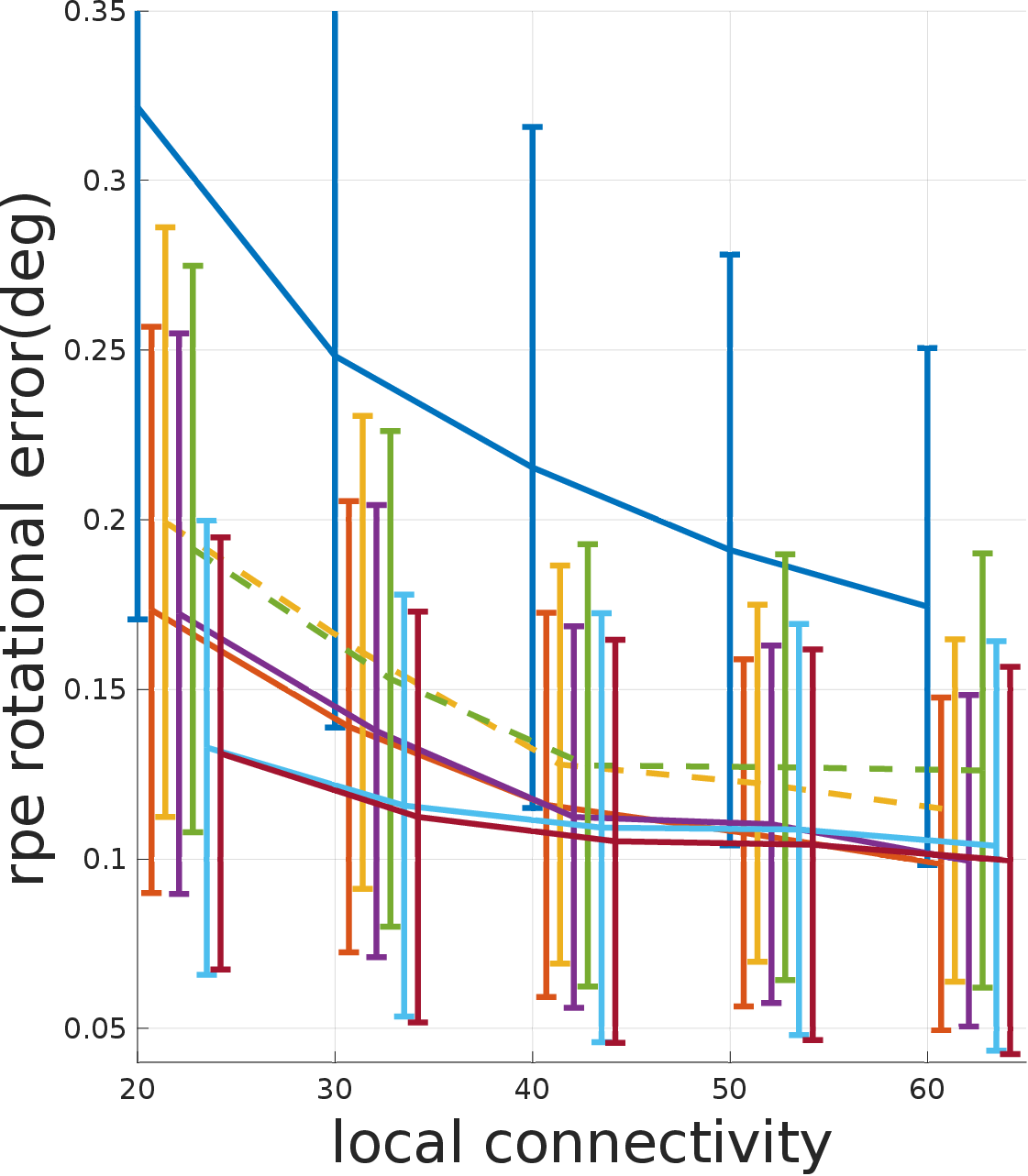}}
\subfigure[]{\label{fig:sim-05-localC-t}\includegraphics[width=0.16\linewidth]{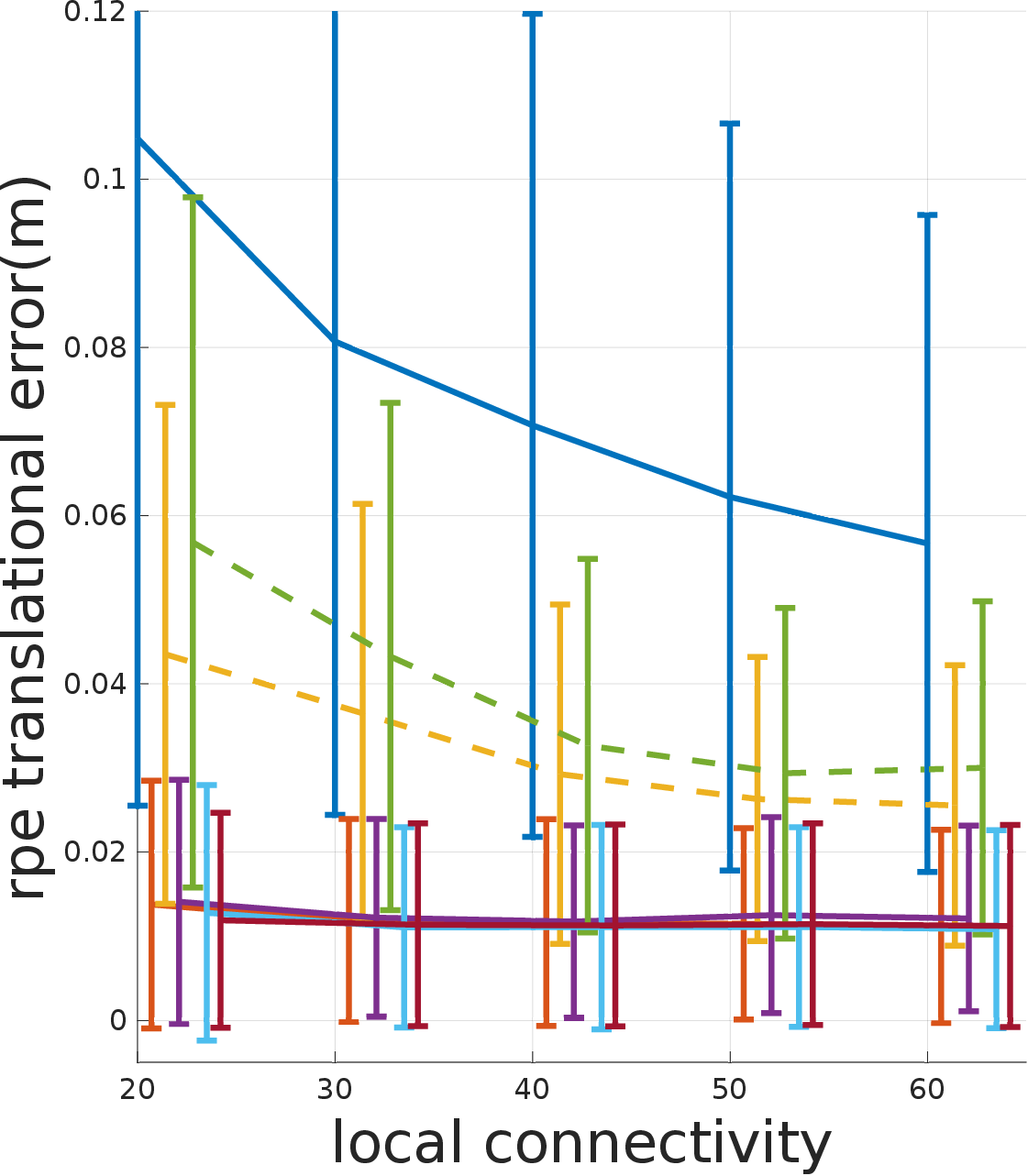}} 
\caption{Mean and standard deviation of RPE for different methods on synthetic data. The default noise level used in the experiments is 4, the default global connectivity is 3, and the default local connectivity is 40. The first row uses KITTI-VO-05 ground truth trajectories, while the second row summarizes results obtained on more sparsely sampled artificial trajectories. Columns one and two analyse rotational and translational errors for varying noise levels, columns 3 and 4 for varying maximum number of observations for each landmark, and columns 5 and 6 for varying number of landmarks per frame.}
\label{fig:sim}
\end{figure*}

Results are indicated in the first row of Figure \ref{fig:sim}. The detailed settings and resulting observations are as follows:
\begin{itemize}
\item \textbf{Noise level}: The noise level is controlled by setting the standard deviation of the normally distributed noise in unit pixels. As shown in Figures \ref{fig:sim-05-noise-R} and \ref{fig:sim-05-noise-t}, adding kinematic constraints leads to a large reduction of errors; the proposed methods using continuous-time parametrizations perform better than CBA in most cases, especially in terms of the translational error. Although CBASpRv and CBARt are generally less stable, they perform best in low noise scenarios. FSBA and SSBARv in turn present high robustness against increasing noise levels.

\item \textbf{Global connectivity}: As shown in Figures \ref{fig:sim-05-globalC-R} and \ref{fig:sim-05-globalC-t}, the proposed kinematically consistent methods perform significantly better than their alternatives as the graph's global connectivity degrades. CBASpRv and CBARt again perform best for high connectivity, though also CBA is competitive at that setting.

\item \textbf{Local connectivity}: As shown in Figures \ref{fig:sim-05-localC-R} and \ref{fig:sim-05-localC-t}, the proposed kinematic methods perform significantly better as the number of observations per frame decreases, with SSBARv and FSBA outperforming other methods.
\end{itemize}

\subsubsection{Comparison against ORB-SLAM}
To conclude, we let our kinematically constrained optimization compete against an established alternative from the open-source community: ORB-SLAM \cite{murORB2}. RPE results are again indicated in Table \ref{tab:orb}.

The results confirm that simple CBA is not able to compete with ORB-SLAM, while methods that impose kinematic constraints return comparable results and occasionally even outperform ORB-SLAM. We would like to emphasise that---although ORB-SLAM also uses CBA in the back-end---it is a heavily engineered framework that performs additional tasks to reinforce the health and quality of the underlying graphical model, while the back-end optimizer in the present ablation study simply uses the map initialized by the front-end solver. We therefore again conclude that the addition of kinematic constraints generally models the motion well, and increases the ability to handle degrading visual measurements.

\tabcolsep=0.11cm
\begin{table}[htb]
\vspace{-0.3cm}
\small
\caption{Comparison against ORB-SLAM. Error in $\mathbf{t}$: [m] and $\mathbf{R}$: [deg].}
\vspace{-0.2cm}
    \centering
    \begin{tabular}{|c|l|c|c|c|c|}
    \hline
    Dataset & method & mean($\mathbf{t}$) & stddev($\mathbf{t}$) & mean($\mathbf{R}$) & stddev($\mathbf{R}$) \\ \hline
    \multirow{5}{*}{VO-01} & ORB-SLAM & 0.1293 & 0.1676 & \textbf{0.3149} & 0.4548 \\
                           & CBA & 0.0170 & 0.0413 & 0.3580 & 0.5445 \\
                           & CBASpRv  & 0.0082 & 0.0046 & 0.3929 & 0.4717 \\
                           & SSBARv   & \textbf{0.0078} & \textbf{0.0033} & 0.3606 & \textbf{0.3684} \\
                           & FSBA     & 0.0080 & 0.0035 & 0.3711 & 0.4131 \\ \hline
    \multirow{5}{*}{VO-04} & ORB-SLAM & 0.0073 & 0.0034 & \textbf{0.0451} & \textbf{0.0312} \\
                           & CBA & 0.0079 & 0.0039 & 0.0775 & 0.0392 \\
                           & CBASpRv  & \textbf{0.0050} & 0.0032 & 0.0784 & 0.0392 \\
                           & SSBARv   & 0.0050 & \textbf{0.0032} & 0.0806 & 0.0419 \\
                           & FSBA     & 0.0051 & 0.0032 & 0.0829 & 0.0435 \\ \hline
    \multirow{5}{*}{VO-06} & ORB-SLAM & 0.0076 & 0.0074 & \textbf{0.0432} & \textbf{0.0277} \\
                           & CBA & 0.0145 & 0.0411 & 0.1039 & 0.2494 \\
                           & CBASpRv  & \textbf{0.0057} & \textbf{0.0065} & 0.0951 & 0.0821 \\
                           & SSBARv   & 0.0057 & 0.0066 & 0.1013 & 0.0779 \\
                           & FSBA     & 0.0058 & 0.0068 & 0.1074 & 0.0791 \\ \hline
    \end{tabular}
    \label{tab:orb}
    \vspace{-0.3cm}
\end{table}

\begin{figure}[htbp]
\vspace{-0.2cm}
\centering
\subfigure[]{\includegraphics[width=0.32\linewidth]{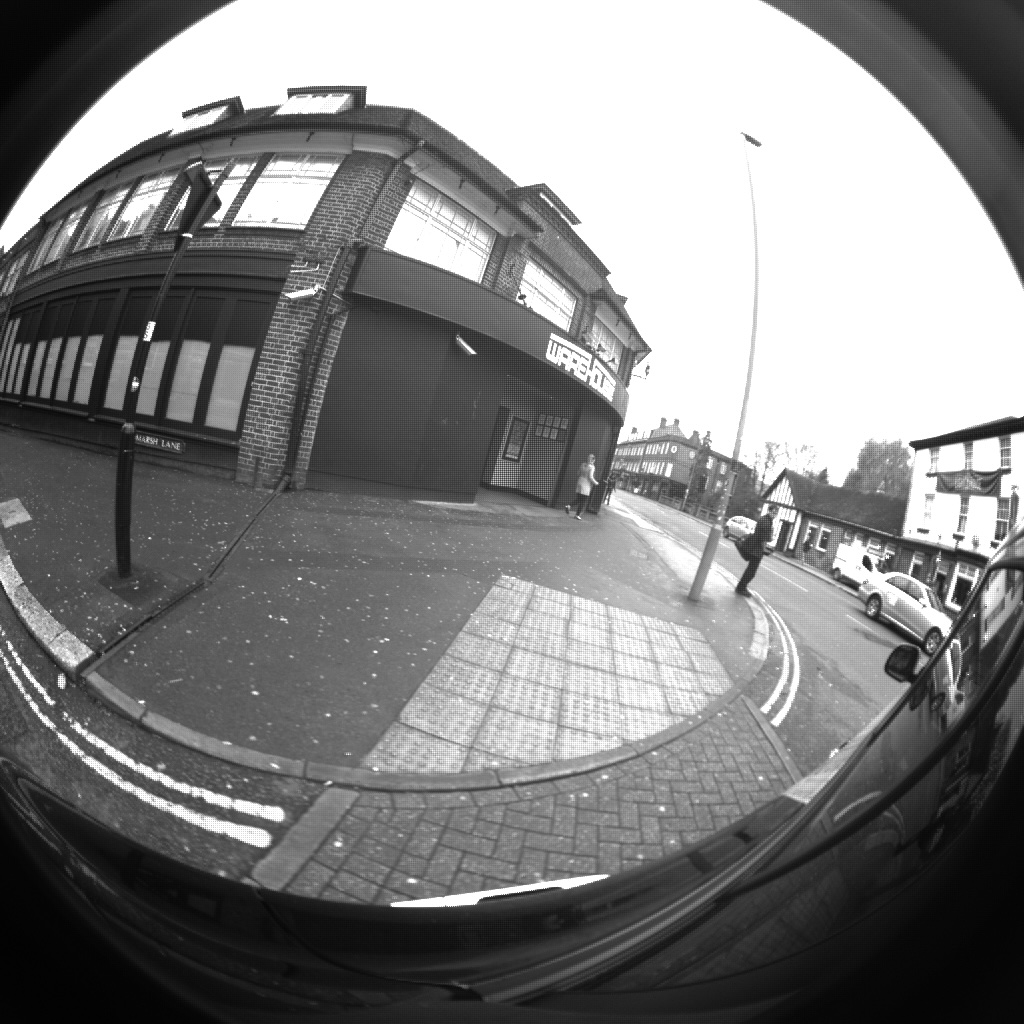}}
\hfill
\subfigure[]{\includegraphics[width=0.32\linewidth]{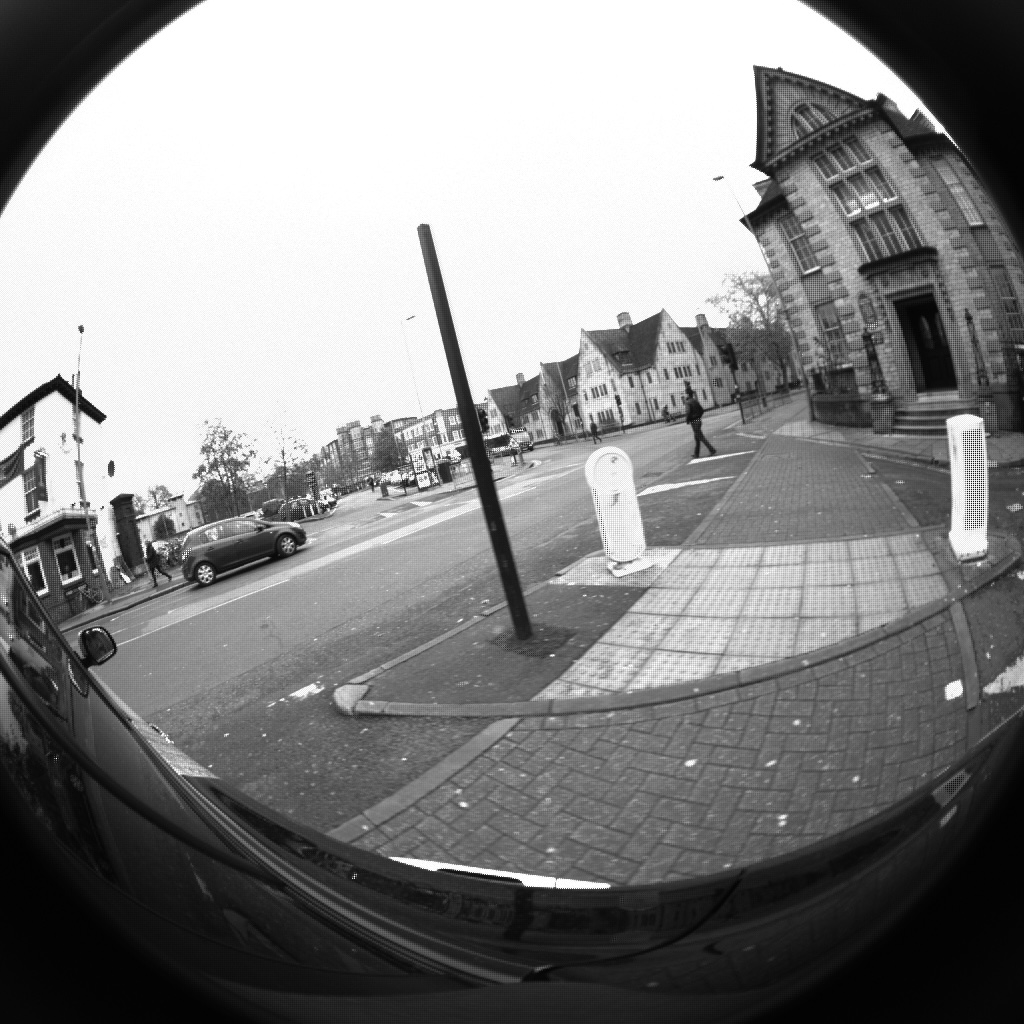}}
\hfill
\subfigure[]{\includegraphics[width=0.32\linewidth]{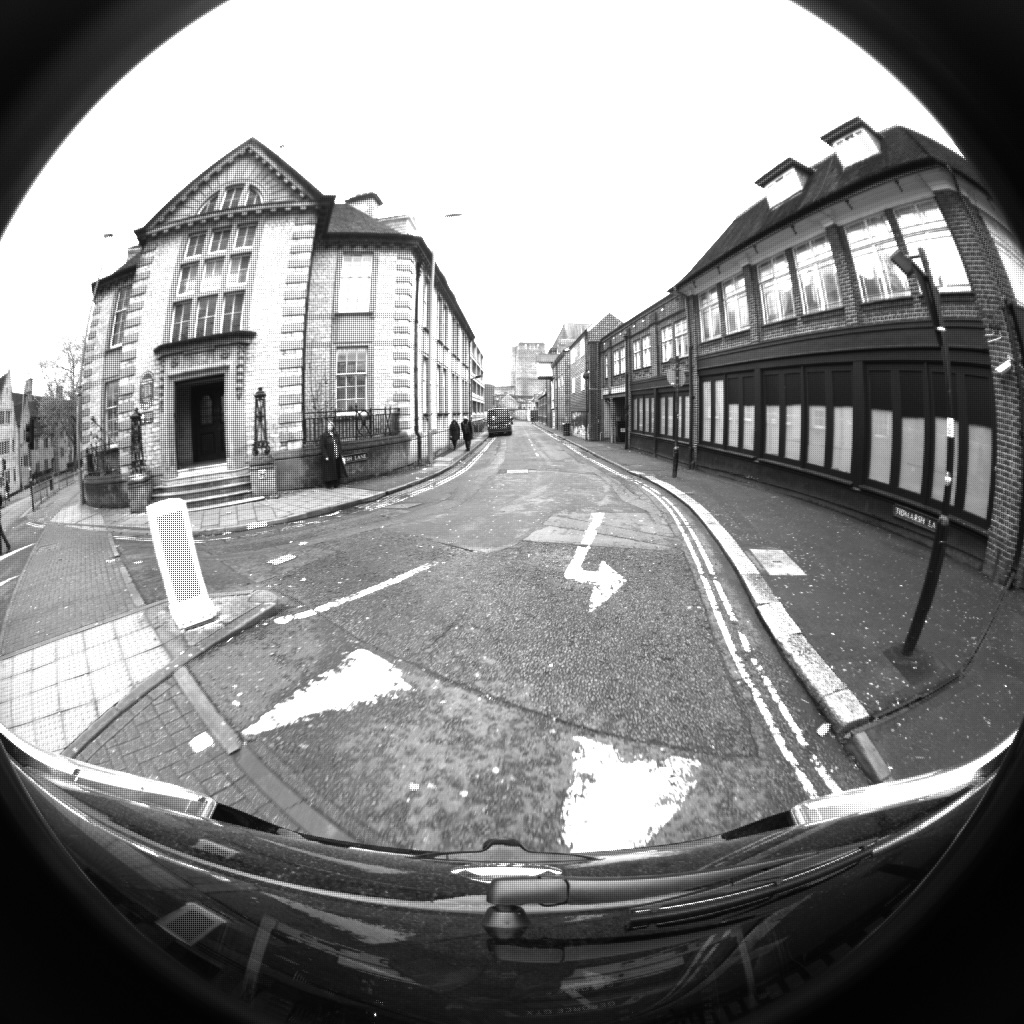}}
\\ \vspace{-0.2cm}
\subfigure[]{\includegraphics[width=0.32\linewidth]{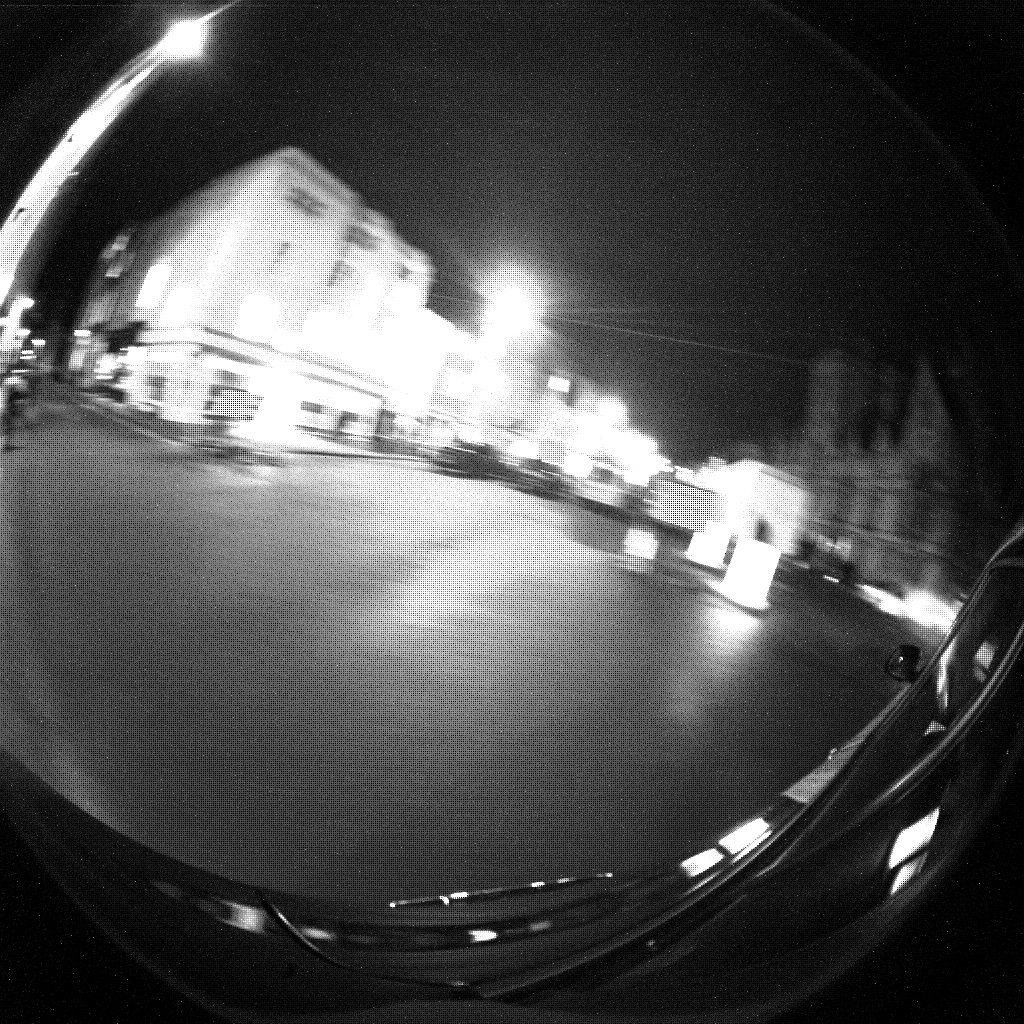}}
\hfill
\subfigure[]{\includegraphics[width=0.32\linewidth]{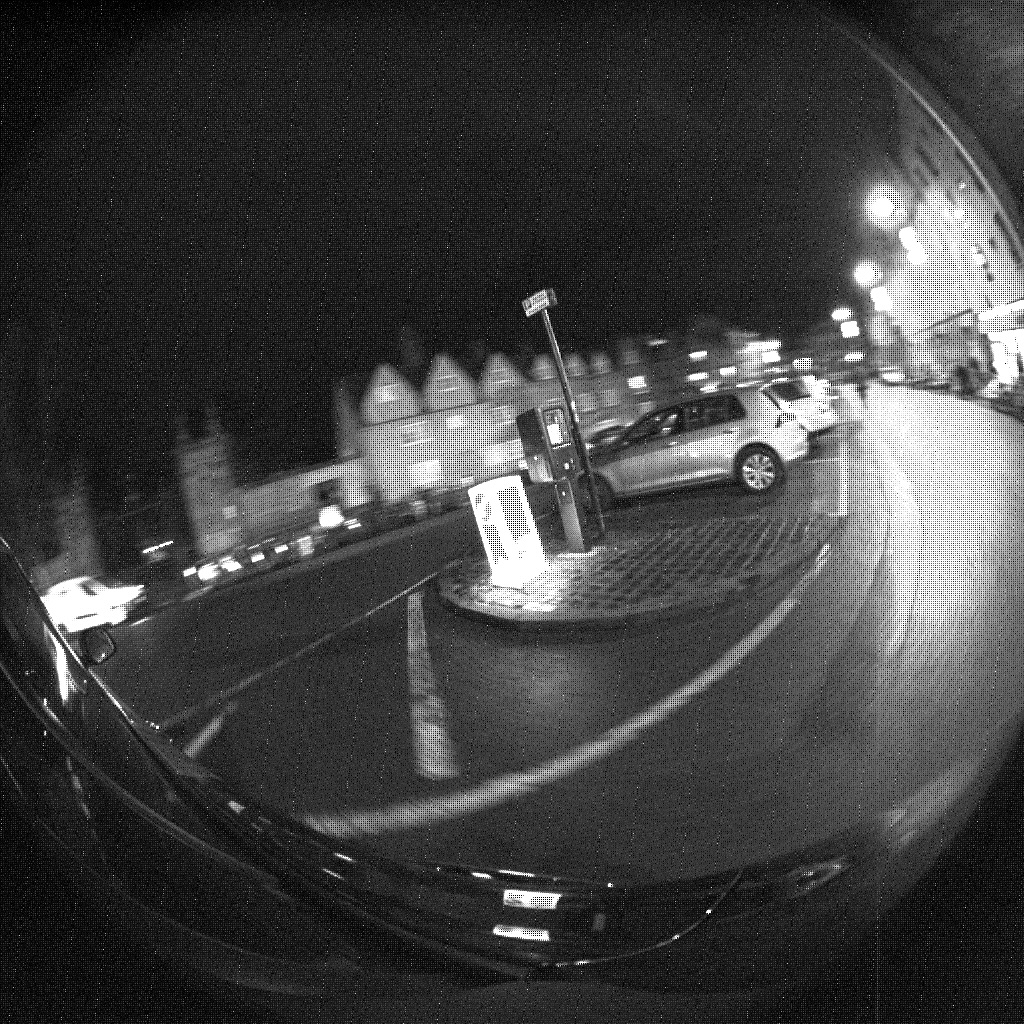}}
\hfill
\subfigure[]{\includegraphics[width=0.32\linewidth]{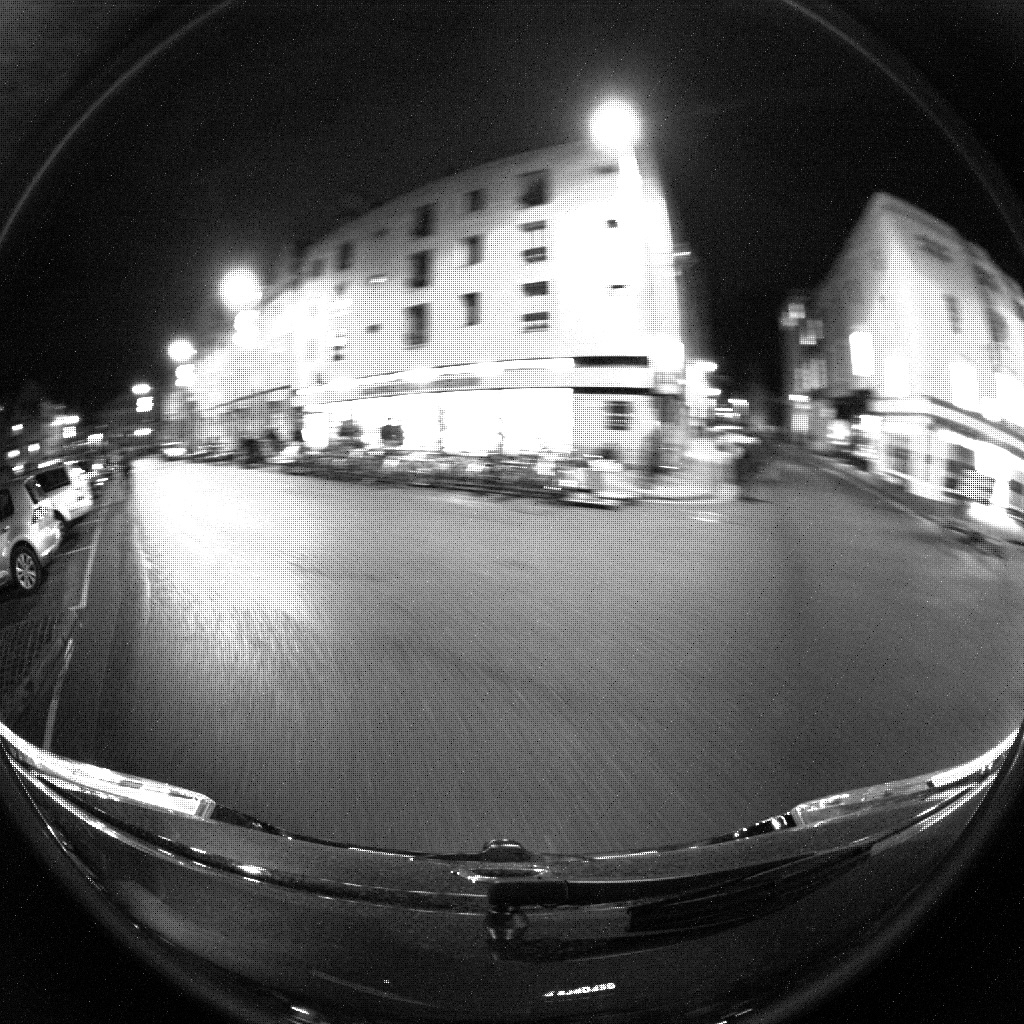}}

\vspace{-0.2cm}
\caption{Sample images of selected sequences from the Oxford RobotCar Dataset~\cite{RobotCarDatasetIJRR}. Images (a) to (c) show views from the left, right, and rear cameras for Sequence 1 (daytime). Images (d) to (f) show views for Sequence 2 (nighttime).}
\vspace{-0.2cm}
\end{figure}

\subsection{Evaluation of complete framework on large-scale scenes}

To evaluate the performance of the complete SLAM system, we conducted large-scale experiments using the Oxford RobotCar Dataset~\cite{RobotCarDatasetIJRR}. This dataset is captured in a challenging environment with over 100 repetitions of a consistent route through Oxford, UK, captured over a year under varying conditions such as weather, traffic density, and pedestrian activity. These variations present an ideal testing ground for assessing the robustness and accuracy of the proposed SLAM system.

\subsubsection{System Configuration} Our framework processes the incoming sparse feature correspondences using the newly proposed OpenGV 2.0 modules for stable motion initialization and spline-based back-end optimization. Although the datasets include a forward-facing stereo camera and three mono cameras pointing towards the sides and the rear of the car, we restrain ourselves to the monocular cameras to demonstrate the system’s capability in handling less ideal setups with very limited overlap between fields of view. Note that for the optional GPS signal, we do not use the kinematic groundtruth presented in~\cite{RCDRTKArXiv}, which relies on a fusion of post-processed GPS signals, inertial readings, and static GNSS base station recordings. Instead, we rely on the raw GPS signals, which are of limited quality and suffer from occasional outliers. Outliers in low-grade GPS receivers are common in urban environments and may be caused by an insufficient number of satellites or multi-path effects.

\subsubsection{Data Selection} We select four sequences that exhibit a wide range of motion patterns and environmental conditions. Specifically, we use:
\begin{itemize}
    \item \textbf{Sequence 1 (2014-11-28-12-07-13):} This sequence is 6.6 kilometers long and was captured in daylight conditions. However, it exhibits significant GPS signal jumps, posing a challenge for trajectory estimation.

    \item \textbf{Sequence 2 (2014-12-16-18-44-24):} Covering 8.4 kilometers, this sequence was recorded in low-light, nighttime conditions. The reduced ambient lighting results in fewer detectable features for visual-based systems, particularly in areas lacking artificial light sources.

    \item \textbf{Sequence 3 (2014-11-18-13-20-12):} This 9.8-kilometer sequence was recorded in daylight under overcast conditions. The diffuse lighting reduces harsh shadows, aiding feature extraction compared to the HDR challenges posed by bright, sunny conditions.

    \item \textbf{Sequence 4 (2015-02-03-08-45-10):} This sequence, recorded over a span of 13.9 kilometers, captures a winter landscape with significant snow coverage. Additionally, GPS signal interruptions are present throughout the sequence, further complicating accurate localization and trajectory estimation.
\end{itemize}

\subsubsection{Evaluation} We evaluate the system in three different configurations:
\begin{itemize}
    \item \textbf{Front-End Only:} Motion initialization and simple scale propagation over successive pairs of multi-perspective view-points within RANSAC.
    \item \textbf{Standard Bundle Adjustment:} The front-end appended by a standard bundle adjustment back-end for pre-optimization.
    \item \textbf{FSBA Spline Optimization:} Result of the complete framework including our proposed FSBA spline-based optimization to continuously enforce non-holonomic constraints and improve trajectory smoothness. The result also includes weak GPS priors.
\end{itemize}

\begin{figure*}[htbp]
\vspace{0.2cm}
\centering
\subfigure[]{\label{fig:sat1_all}\includegraphics[width=0.32\linewidth]{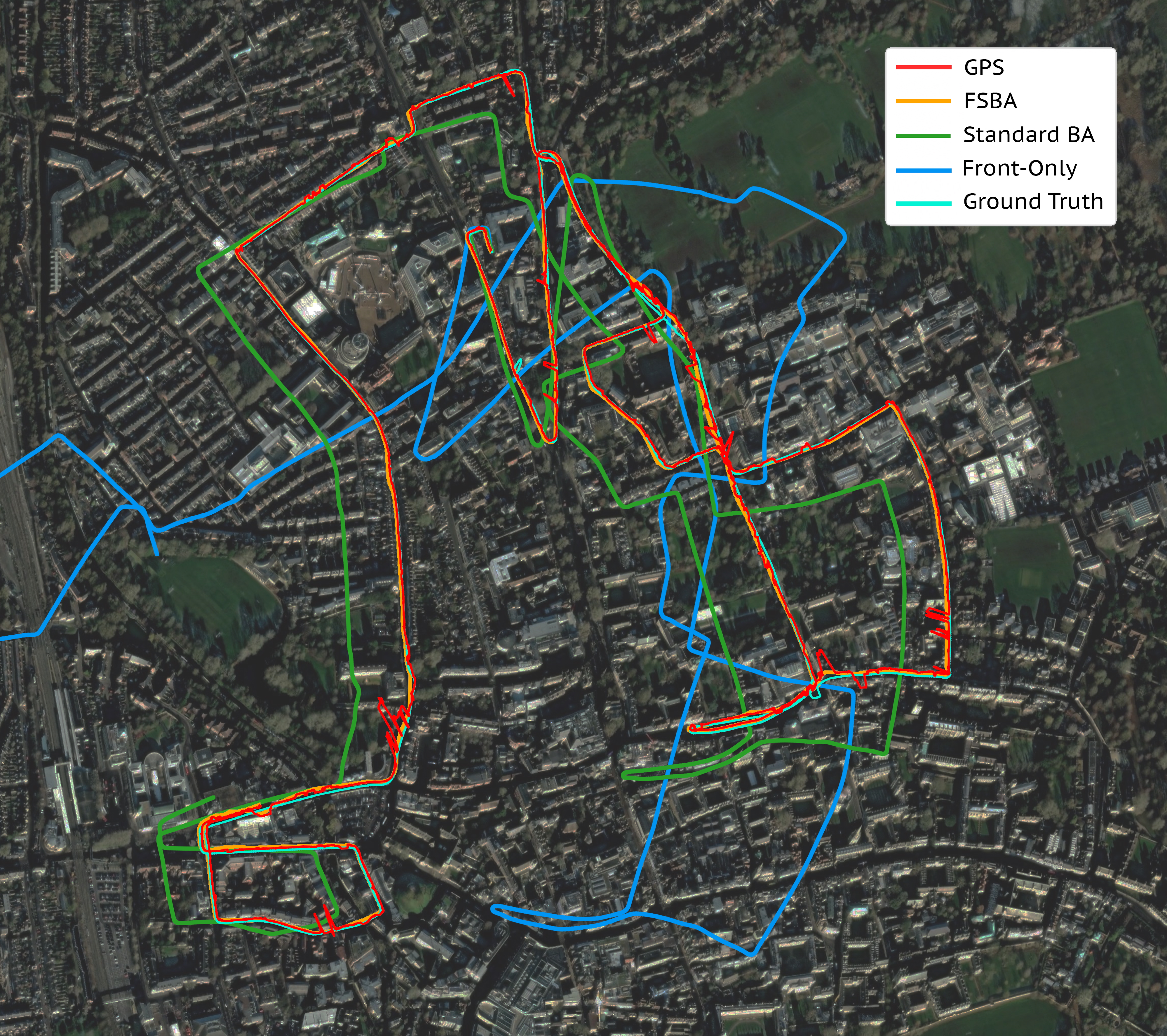}} \hfill
\subfigure[]{\label{fig:sat2_all}\includegraphics[width=0.32\linewidth]{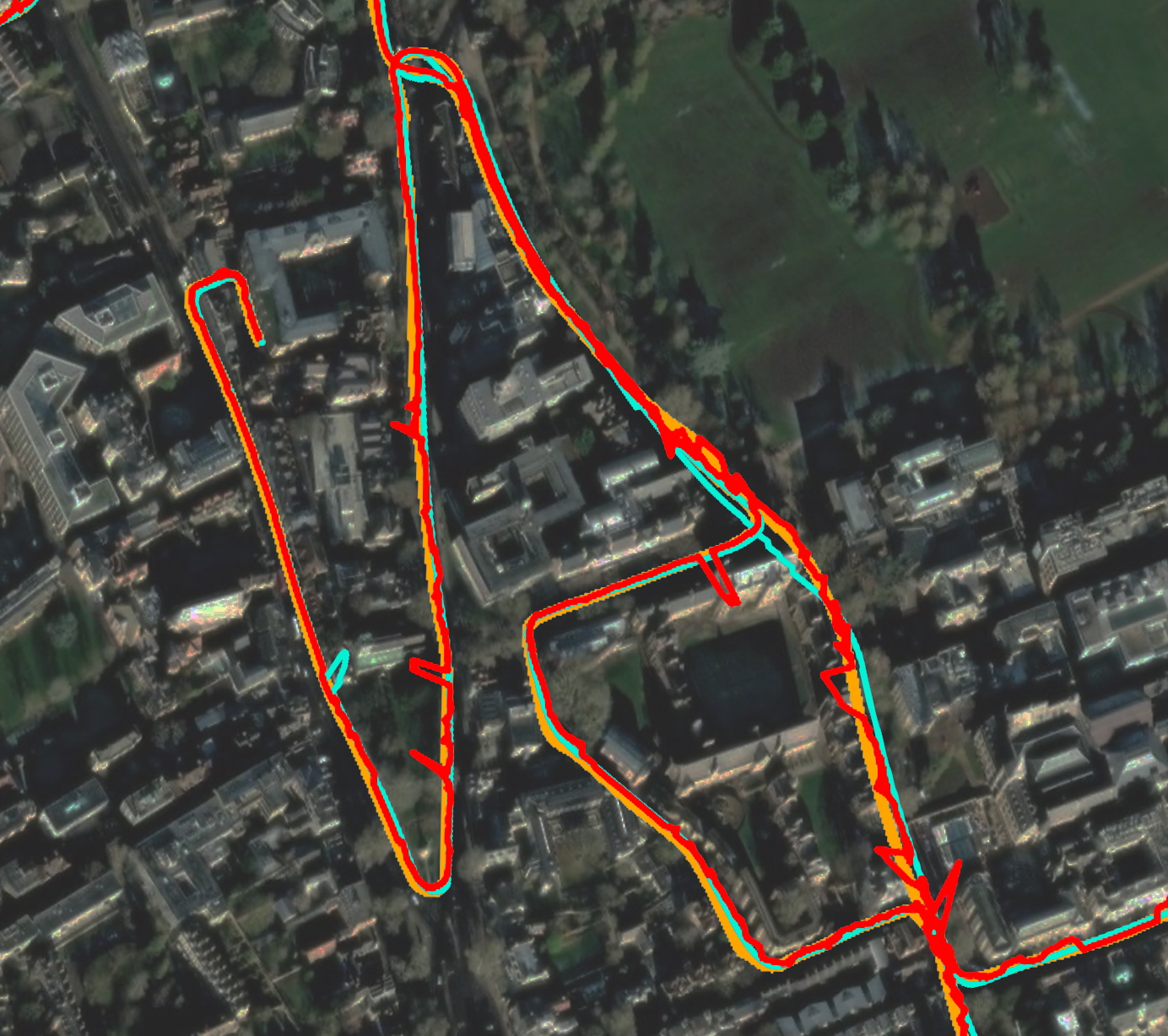}} \hfill
\subfigure[]{\label{fig:sat1_det}\includegraphics[width=0.32\linewidth]{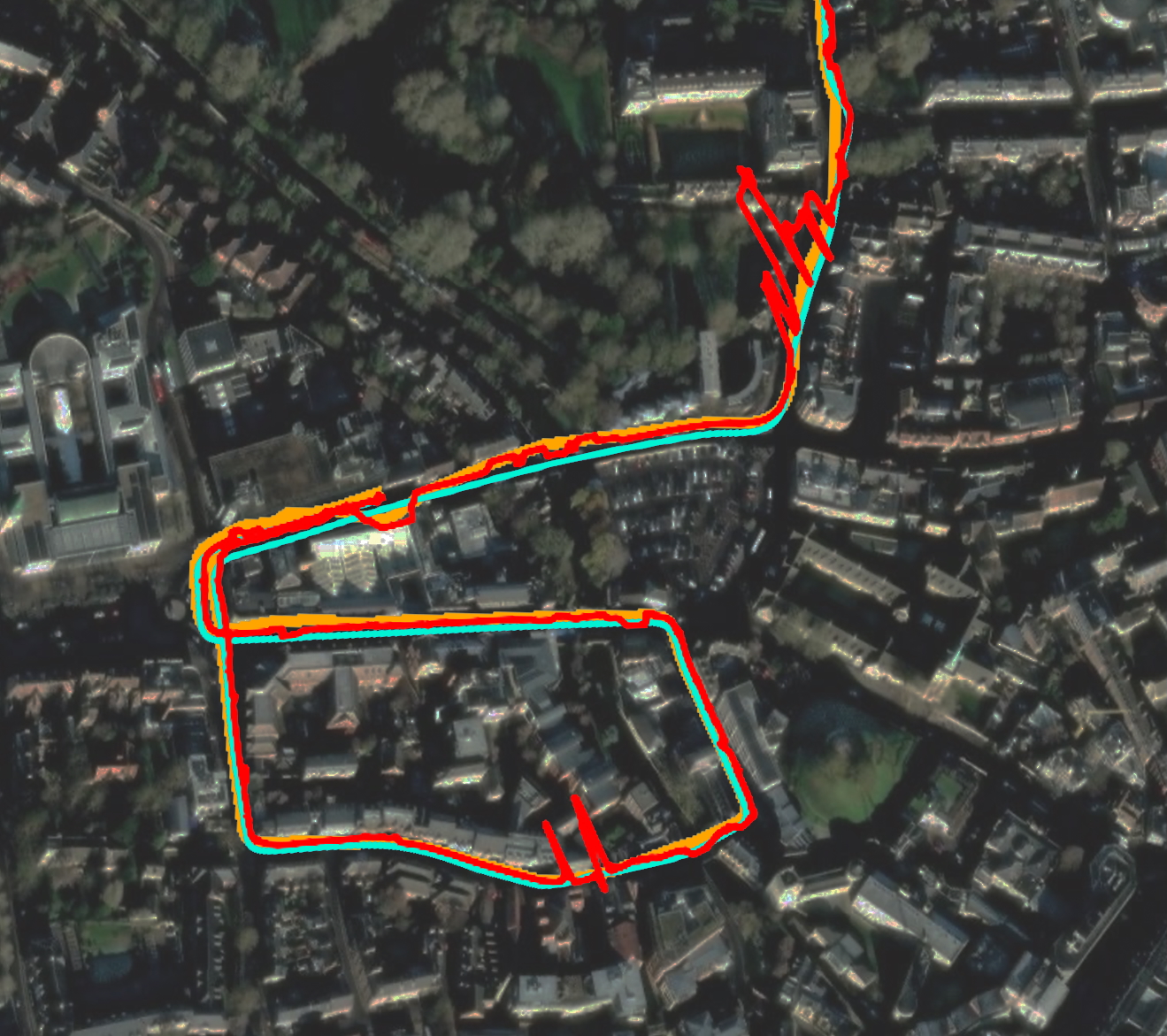}}
\vspace{-0.2cm}
\caption{Trajectories of the SLAM system plotted on a satellite map for the selected sequences from the Oxford RobotCar Dataset~\cite{RobotCarDatasetIJRR}. (a) is the trajectory for Sequence 1. As can be observed, our proposed FSBA with weak GPS data terms naturally returns the best result, and improves over the front-end or the result coming out of standard bundle adjustment, even better than groundtruth from~\cite{RCDRTKArXiv} in some parts. As shown in (b) and (c) --- detailed views of Sequence 1 --- the GPS signal is at times unreliable. Our method successfully detects and ignores faulty readings all while ensuring a kinematically feasible trajectory.}
\vspace{-0.5cm}
\end{figure*}

\subsubsection{Results} Results for sequence 1 are shown in  Figure~\ref{fig:sat1_all}. The comparative trajectory estimation results obtained for the different optimization methods is overlaid on a satellite view of the environment. As can be observed, the front-end only method suffers from significant drift accumulation owing to the lack of multi-view constraints, which leads to fast drift accumulation. Standard bundle adjustment leads to a significant reduction of drift and improves the alignment, but still fails to perfectly align with the street view due to pose drift. Finally, the proposed FSBA method complemented by a careful inclusion of weak GPS priors leads to the best trajectory estimation results as it corrects GPS jumps while ensuring a smooth, accurate, and kinematically feasible path. As visualized in the detailed view shown in \ref{fig:sat1_det}, our kinematics-aware method maintains a consistent trajectory even in segments where the GPS signal is unreliable.

Sequence 2 presents the most challenging scenario. It consists of a low-light scenario, where the system faces challenges posed by poor feature quantity and quality. The front-end only method again fails to produce an accurate trajectory and suffers from significant drift. Compared against the daylight scenario, standard bundle adjustment in low-light conditions also returns trajectory estimates that are affected by increased drift accumulation. Finally, FSBA with weak GPS priors manages to maintain an accurate trajectory. Note however that at some point in the sequence the number of extracted correspondences falls below a hard minimum required for stable front-end execution, hence not the full dataset is processed.

For Sequence 3, which was recorded under overcast daylight conditions, FSBA demonstrates a significant improvement over standard BA. The reduced HDR effects in overcast conditions allow for more consistent feature extraction compared to sequences recorded in brighter conditions, where HDR challenges can degrade feature matching. While FSBA achieves a lower error compared to standard BA, it still faces difficulties aligning the full trajectory, particularly due to some challenging areas with low-texture surfaces and sharp turns that reduce the effectiveness of feature-based visual odometry. The use of weak GPS priors in FSBA+GPS helps to mitigate these issues.

Sequence 4 is significantly longer than the other sequences, spanning 13.9 kilometers. Despite this, standard BA performs similarly to Sequence 3 in terms of RMSE and APE. Although the snow creates a different visual environment, it doesn't cover many features, making the scenario comparable to the overcast conditions in Sequence 3, where the reduced lighting variability helps feature extraction. FSBA manages to improve over standard BA, but the performance is similar to Sequence 3. The inclusion of weak GPS priors in FSBA+GPS further reduces the APE, underscoring the advantage of using GPS data to refine trajectory estimates over longer distances and under consistent visual conditions.

Numerical results comparing to groundtruth ~\cite{RCDRTKArXiv} are presented in Table \ref{tab:multivo}.

\tabcolsep=0.11cm
\begin{table}[htb]
\vspace{-0.3cm}
\small
\caption{Comparison against groundtruth. Error in [m].}
\vspace{-0.2cm}
    \centering
    \begin{tabular}{|c|l|c|c|c|c|}
    \hline
     & method & $\text{RPE}_\text{median}$ & $\text{RPE}_\text{rmse}$ & $\text{APE}_\text{median}$ & $\text{APE}_\text{rmse}$ \\ \hline
    \multirow{4}{*}{Seq 1} 
    & GPS      & 0.908260 & 6.355607 & 5.438066 & 24.083009 \\
    & Std BA   & 0.114031 & 0.393072 & 91.992155 & 120.381067 \\
    & FSBA     & 0.137423 & \textbf{0.326964} & 52.444591 & 66.039714 \\
    & FSBA+GPS & \textbf{0.079175} & 0.687518 & \textbf{2.298782} & \textbf{8.136368} \\ \hline
    \multirow{4}{*}{Seq 2} 
    & GPS      & 1.230241 & 3.058468 & 2.262350 & 5.098881 \\
    & Std BA   & 0.071944 & 0.147210 & 95.842793 & 128.803591 \\
    & FSBA     & \textbf{0.071351} & \textbf{0.141368} & 52.444591 & 66.039714 \\
    & FSBA+GPS & 0.074188 & 0.432637 & \textbf{2.057649} & \textbf{4.168152} \\ \hline
    \multirow{4}{*}{Seq 3} 
    & GPS      & 1.045106 & 3.168914 & 2.779416 & 5.713864 \\
    & Std BA   & 0.082383 & 0.599251 & 193.315248 & 217.057791 \\
    & FSBA     & 0.110620 & \textbf{0.286018} & 112.841528 & 125.721049 \\
    & FSBA+GPS & \textbf{0.068255} & 0.328448 & \textbf{2.511050} & \textbf{3.952005} \\ \hline
    \multirow{4}{*}{Seq 4} 
    & GPS      & 1.026643 & 3.527595 & 2.210903 & 5.843370 \\
    & Std BA   & 0.079679 & 0.296779 & 182.068247 & 196.111795 \\
    & FSBA     & 0.643875 & \textbf{0.098930} & 98.525997 & 150.140482 \\
    & FSBA+GPS & \textbf{0.057054} & 0.368412 & \textbf{2.046866} & \textbf{5.0385981} \\ \hline
    \end{tabular}
    \label{tab:multivo}
    \vspace{-0.3cm}
\end{table}

In general, the obtained experimental results demonstrate the stability and accuracy of the system under various challenging conditions. The addressed challenges include stop-and-go motion in a vibrant urban environment, significant GPS jumps, and low-light conditions. Our proposed FSBA spline-based optimization method consistently produces high localization accuracy, demonstrating the integration of non-holonomic motion constraints into a SLAM framework significantly enhances the stability and fidelity of the estimated trajectories. To the best of our knowledge, our result marks the first time that a SLAM system that primarily relies on surround-view, non-overlapping visual inputs has been successfully used to consistently produce large-scale trajectory estimation results. The proposed system is also attractive in terms of computational efficiency, and enables real-time processing at 11 surround-view captures per second on a standard laptop. We believe this number to be highly encouraging in a large spectrum of potential real-world applications.

\section{Conclusion}
\label{sec:conclusion}

The present paper introduces motion prior-supported calibration and motion estimation solutions for pure non-overlapping multi-perspective cameras, thereby enabling a new level of performance for challenging surround-view camera systems with severely reduced overlap between neighbouring fields of view. We have proposed a new extrinsic orientation optimization paradigm that operates on natural data, thereby providing a practical solution to the refinement of default or factory calibration parameters, or an effective remedy to account for changes over time as vibrations or shocks occur. We have furthermore introduced a novel planar displacement solver for multi-camera systems which stays robust irrespective of the specifics of the camera motion. It does not assume the more approximative circular arc model, and presents no scale unobservabilities in degenerate motion conditions. This is important as the degenerate conditions for existing algorithms are often closely related to common vehicle displacements (e.g. pure translational displacements). To conclude, we have introduced a spline-based parameterization for the temporal vehicle pose that continuously enforces the kinematic constraints given on common ground vehicles, and thereby robustifies and improves the results obtainable from non-overlapping surround-view camera systems. Note that none of the proposed methods makes the circular arc assumption, which is indeed valid on an instantaneous level, only. Furthermore, even if slippage occurs, non-holonomic models relying on an instantaneous centre of rotation still explain the motion of skid-steering platforms relatively well~\cite{martinez05}, which is why the proposed theory may also be applied to such platforms. Our contribution is rounded off by an inclusion into an open source framework that handles all particularities of vehicle motion in urban environments and takes care of metric scale observability and drift accumulation via the inclusion of weak GPS priors. Our framework will be open-source released once accepted.

\bibliographystyle{IEEEtran}
\bibliography{opengv2.bib}

\end{document}